\documentclass[10pt,twocolumn,letterpaper]{article}

\usepackage{3dv}
\usepackage{times}
\usepackage{epsfig}
\usepackage{graphicx}
\usepackage{amsmath}
\usepackage{amssymb}

 \graphicspath{{figure/}}
\usepackage{graphicx} 
\usepackage{enumitem}
\usepackage{caption}
\usepackage{dsfont}
\usepackage{amsfonts,amsmath,amssymb,amsthm}
\usepackage{bm,nicefrac}

\usepackage[ruled,vlined,linesnumbered]{algorithm2e}
\usepackage[dvipsnames,svgnames,x11names]{xcolor}
\usepackage{algorithmicx}
\usepackage{algpseudocode}

\usepackage[labelfont=bf, labelsep=period]{caption} 

\usepackage[pagebackref=true,breaklinks=true,letterpaper=true,colorlinks,bookmarks=false]{hyperref}

\threedvfinalcopy 

\def\threedvPaperID{****} 

\begin{document}

\title{Neural 3D Clothes Retargeting from a Single Image}

\author{
Jae Shin Yoon$^\dagger$
\hspace{10mm}Kihwan Kim$^\sharp$
\hspace{10mm}Jan Kautz$^\sharp$
\hspace{10mm}Hyun Soo Park$^\dagger$
\\
\hspace{-10mm}$^\dagger$University of Minnesota
\hspace{25mm}
$^\sharp$NVIDIA \\
}
\twocolumn[{%
\maketitle
\vspace{-12mm}
	\begin{center}
		\centering
	\includegraphics[width=7 in]{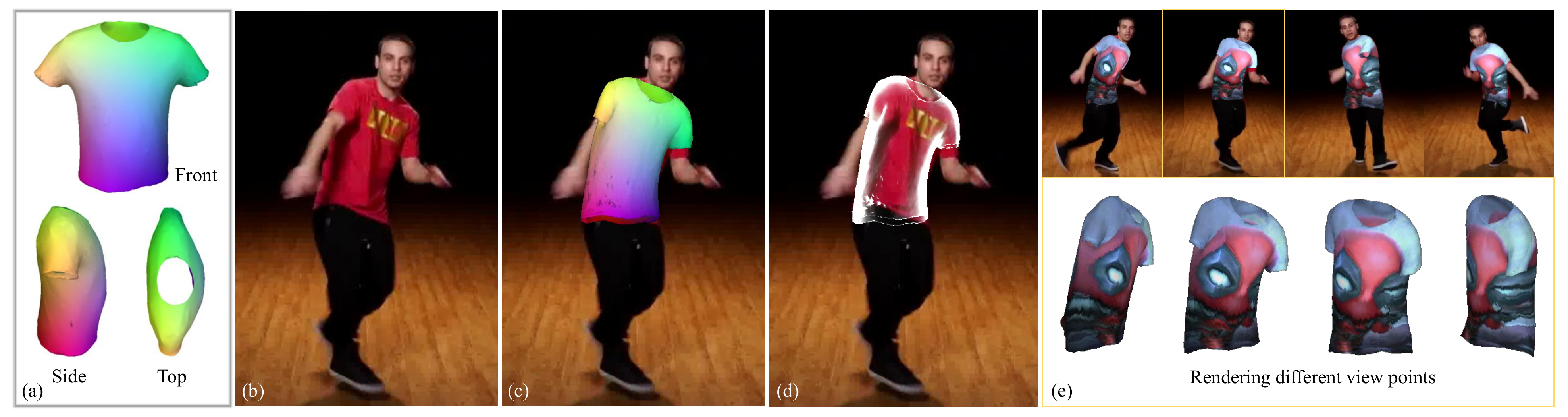}
	\vspace{-7mm}
	\captionof{figure}{Given (a) a reference 3D clothing model and (b) a single 2D image with a target person, (c) we \textit{retarget} the model onto the image by predicting the most plausible cloth deformation and camera pose. (d) The shaded mesh shows that the retargeted 3D clothing is well-aligned with the one in the input 2D image. (e) We demonstrate the retargeting to the arbitrary poses of the same person with a texture.
	}
	\label{fig:teaser}
\end{center}	
\label{fig:teaser_main}
}]

\begin{abstract}
In this paper, we present a method of clothes retargeting; generating the potential poses and deformations of a given 3D clothing template model to fit onto a person in a single RGB image. The problem is fundamentally ill-posed as attaining the ground truth data is impossible, i.e., images of people wearing the different 3D clothing template model at exact same pose. We address this challenge by utilizing large-scale synthetic data generated from physical simulation, allowing us to map 2D dense body pose to 3D clothing deformation. With the simulated data, we propose a semi-supervised learning framework that validates the physical plausibility of the 3D deformation by matching with the prescribed body-to-cloth contact points and clothing silhouette to fit onto the unlabeled real images. A new neural clothes retargeting network (CRNet) is designed to integrate the semi-supervised retargeting task in an end-to-end fashion. In our evaluation, we show that our method can predict the realistic 3D pose and deformation field needed for retargeting clothes models in real-world examples.
\vspace{-.4cm}
\end{abstract}



\section{Introduction}

This paper studies the problem of {\em clothes retargeting}---generating the plausible 3D deformation of a given clothing template to fit onto people with arbitrary poses and appearances from a single RGB image. Clothes retargeting brings out a novel capability of clothes reenactment in a photorealistic fashion. For instance, in Fig.~\ref{fig:teaser}, given an image of a dancer and a 3D template of a T-shirt model, we predict the deformation of the template clothing that differs from the clothing in the image, allowing rendering of newly texture-mapped appearance, i.e., 3D virtual try-on\footnote{Existing virtual try-on interfaces are by large limited to a certain actor at fixed viewpoint.}.

Beyond the challenges of single view reconstruction, clothes retargeting introduces a new challenge of clothing transfer that requires two types of correspondences; 
(1) Image-to-cloth: to match a 3D clothing to the target image, the geometry of the clothing shown in the image needs to be estimated from a single view image. The existing single view based human body reconstruction~\cite{bogo2016keep,xiang2018monocular} does not apply because there is no parametric model for a cloth unlike a human body model~\cite{loper2015smpl}.
(2) Cloth-to-template: our retargeting aims to transfer the geometric deformation of the clothing in the image to the template clothing regardless of their appearances and topology (i.e., T-shirt onto a long sleeve shirt). Since it is impossible for the same person to wear different clothes with exactly identical pose, attaining cloth-to-template correspondences is not trivial.

We hypothesize that these two challenging correspondence problems can be addressed by embedding the underlying physics of clothing into visual recognition in three ways. (1) Clothing undergoes deformation through physical interactions with the body, which can be simulated with respect to nearly all possible body pose configurations~\cite{kim:2013}. This simulation allows us to further learn the spatial relationship between 2D body shape and 3D template clothing even without knowing the perfect geometry of actual clothing in the image. (2) Global clothing deformation could be expressed by a few contact points on the body, e.g., collar, where the distance between the points on the body and those from clothing roughly remain constant. 
We identify these points on the clothing template, and fit the reconstructed template to the target image. (3) Clothing deformation is also influenced by physical interactions with the environment, such as wind. Such interactions may result in similar geometric deformations, e.g., the clothing silhouette must align with the reconstructed template~\cite{habermann2020deepcap,habermann2019livecap,xu2018monoperfcap}. This could enforce a more plausible fit of the template even when there is a topological gap between the template and the actual clothing in the image.

Based on these hypotheses, we formulate a semi-supervised clothes retargeting method that benefits from both synthetic geometry and real images using a Clothes Retargeting Network (CRNet). CRNet is designed to learn the spatial relationship between the body and a template cloth from synthetic data and is adapted to real images by matching the prescribed body contact points and clothing silhouette. Further, we introduce an online clothing refinement algorithm to adjust the clothing  reconstruction and camera pose for a testing data. In our evaluation, we show that our method can accurately estimate the 3D pose and deformation field needed for retargeting clothing models in real-world examples in the presence of occlusion, motion blur, and multiple instances.

The key strength of our method is that we can learn a garment-specific deformation model in a fully automatic way with physical simulation and adapt this model to real image without any labeled data, allowing us to apply our retargeting pipeline to a wide variety of scenes. In addition, we emphasize that our method can skip to estimate the 3D body pose and directly retarget the 3D clothing from a single image by implicitly parameterizing the body pose prior learned from large-scale synthetic data, which saves much computational time for fitting and rendering the 3D human body. The main contributions of this work include: (1) an end-to-end learning framework for single-view clothes retargeting and its adaptation to real image with self-supervision, (2) A method to synthesize 3D clothing deformation and the associated 2D data.



\section{Related Work}
Our clothes retargeting leverages human pose/shape estimate to compute the underlying geometry of the template clothing. 

\noindent \textbf{Single View Human Reconstruction} Inspired by Johansson's point light display experiment~\cite{johansson:1973}, the spatial relationship of the human body has been actively studied to recover 3D humans~\cite{bregler:2000,ozden:2004,bue:2006,torresani:2008}. For instance, diverse facial expressions can be modeled by a linear combination of blend shape bases~\cite{tuan2017regressing}. 
Such models have been combined with facial landmark recognition approaches using deep learning~\cite{tewari2017mofa,zhu2019face,richardson2017learning,tewari2018self,Yoon_2019_CVPR}, allowing an end-to-end face reconstruction from an image.
Unlike the face, a body undergoes nonlinear articulated motion that cannot be well approximated by linear shape bases. This requires additional spatial constraints to express diverse poses, e.g., a limb length constraint~\cite{park:2011,taylor:2000}, activity specific pose~\cite{ramakrishna:2012eccv}, and a physical constraint (e.g., joint force and torque)~\cite{brubaker:2007,wei:2010}. Recently, a large volume of literature has shown that the nonlinear manifold of human pose space can be effectively approximated by training data~\cite{bogo2016keep,kanazawa2018end,guan2009estimating,lassner2017unite,varol2017learning}. 

\begin{figure*}[t]
	\centering
    \includegraphics[width=.99\linewidth]{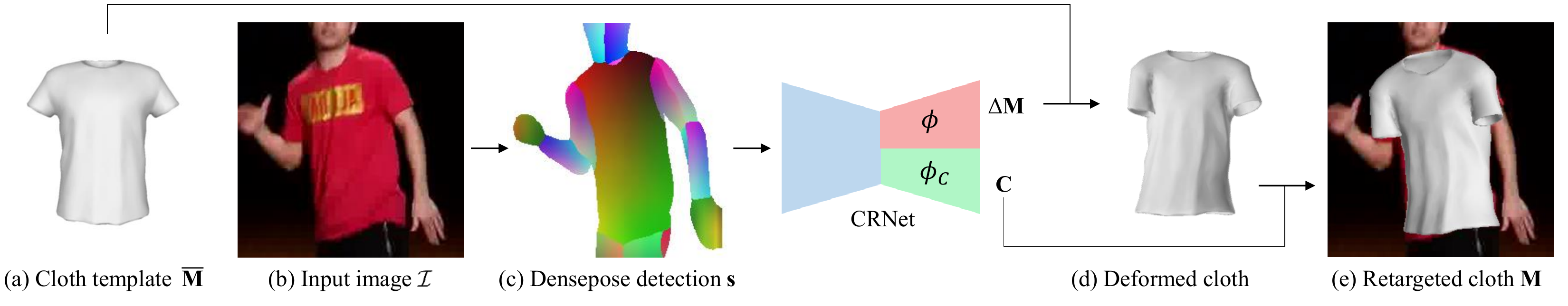}
\caption{\small{\textbf{Overview of our framework}: (a, b) The input to our clothes retargeting framework is a single cropped image ($\mathcal{I}$) with $256\times256$ resolution that contains a person and 3D template clothing mesh $\overline{\mathbf{M}}$. (c) We use 2D human body correspondences ($\mathbf{s}$)~\cite{alp2018densepose} as an initialization for the Clothes Retargeting Network (CRNet), and (d) we then deform $\overline{\mathbf{M}}$ using the predicted deformation field $\Delta \mathbf{M}$. (e) Finally, We overlay the deformed clothing onto $\mathcal{I}$ with the camera pose $\mathbf{C}$. }
}
\label{fig:overview}
\vspace{-.3cm}
\end{figure*}

\noindent \textbf{Clothes Reconstruction} Unlike the face and body, clothes reconstruction is particularly difficult because there is no off-the-shelf model that can represent their deformation. Further, the geometry is commonly very complex, e.g., having wrinkles, and recovering such details requires high resolution 3D measurements such as a multi-camera system~\cite{pons2017clothcap,lahner2018deepwrinkles} or depth camera~\cite{yu2019simulcap,dou2016fusion4d}. Multi-view stereo has been used to recover the high resolution depth and normal maps and to track their deformation~\cite{2008:aguiar,furukawa:2008,joo_cvpr_2014}. RGBD imaging provides an opportunity to measure high resolution cloth at the wrinkle level~\cite{Ye2012KinectsMocap,yu2019simulcap,dou2016fusion4d,Doublefusion,dou:2016}. 

Clothes reconstruction from a single RGB image has been studied in the context of nonrigid structure-from-motion~\cite{torresani:2002,olsen:2007,bartoli:2008,bue:2008} with an assumption that the cloth deformation lies in a latent subspace where shape basis models can be learned online. However, such approaches model the cloth deformation in isolation where the physical interactions with the body surfaces are not taken into account. A key challenge lies in the lack of data that models the precise relationship between body and clothing. Physics simulation is a viable solution that can create extensive amounts of data to learn the relationship, allowing extreme high fidelity reconstruction at wrinkle level~\cite{danvevrek2017deepgarment,gundogdu19garnet,goldenthal2007efficient,li2018implicit}. Since these approaches are built upon physics simulation, there is a domain gap between the data and the real-world cloth deformation. To bridge this gap, multi-view bootstrapping~\cite{varol2017learning}, i.e., training a model with synthetic and real-world multi-view data together, is used, while multi-view data still has the domain gap with real-world data. Unsupervised adversarial loss helps the domain adaptation by adjusting the model feature distribution~\cite{chen2016synthesizing} to the real-world domain, but this adversarial learning tends to be difficult to train without any explicit correspondence. 

\noindent \textbf{Retargeting} With significant commercial interest in facial animation, faces have been the main target to transfer expressions to animated virtual characters (retargeting) or real actors (reenactment). The shape coefficients of facial expressions can be transferred to a target image to synthesize a new image~\cite{vlasic:2005,garrido:2014,thies2016face,Shlizerman:2010}. However, retargeting in the
clothes domain is largely unexplored area. Clothes shape can be transferred across different body shapes and poses~\cite{pons2017clothcap,xu2019denserac,santesteban2019learning} based on cloth-to-body 3D correspondences or manual deformation of a 2D image~\cite{Fadaifard2013}.

In this paper, we aim to retarget a 3D template clothing to the clothes in a single RGB image via underlying geometry. Note that the clothes retargeting differs from garment style transfer techniques~\cite{Raj:2018,Zanfir_2018_CVPR,han:2018,zhu2017be} which is often limited to 2D image synthesis. Zheng et al.~\cite{zheng19try} proposed a clothes retargeting pipeline, which, yet still relies on 2D warping and adversarial training for 2D image synthesis conditioned on an input target pose.
3D clothes reconstruction from a set of images~\cite{Yang:2016,pons2017clothcap,danvevrek2017deepgarment,lahner2018deepwrinkles} also differs from our goal where we seek a deformation transfer to the template clothing that is not necessarily identical to the observed geometry, material, and topology in the target image.




\section{Approach}
\label{sec:approach}

Consider a template clothing mesh $\overline{\mathbf{M}} \in \mathds{R}^{M\times 3}$ with the canonical body pose (T-pose) $\overline{\mathbf{S}} \in \mathds{R}^{M_s\times 3}$ that stores the 3D coordinates of the vertices where $M$ and $M_s$ are the number of vertices in the clothing and body (e.g., SMPL body model~\cite{loper2015smpl}), respectively. We model the deformed clothing as a function of pose and shape of the body:
\begin{align}
    \mathbf{M} = \overline{\mathbf{M}} + f(\mathbf{S}), \label{Eq:deform}
\end{align}
where $f:\mathds{R}^{M_s\times 3}\rightarrow \mathds{R}^{M\times 3}$ maps body pose and shape to the 3D clothing deformation, i.e., $f(\overline{\mathbf{S}}) = \mathbf{0}$ and $\mathbf{S}$ denotes the deformed body model. We denote the deformation as $\Delta \mathbf{M}=f(\mathbf{S})$. The representation of the deformation in Equation~(\ref{Eq:deform}) significantly differs from that of the face and body~\cite{booth2018large,loper2015smpl} in two ways. First, the shape and pose are no longer linear, e.g., a set of linear blend shape bases has been used to model their deformation. Second, the deformation is conditioned on the underlying body shape and pose $\mathbf{S}$, i.e., it is a secondary deformation where the clothing and body need to be learned jointly.

\begin{figure*}[t]
	\begin{center}
\includegraphics[width=0.8\linewidth]{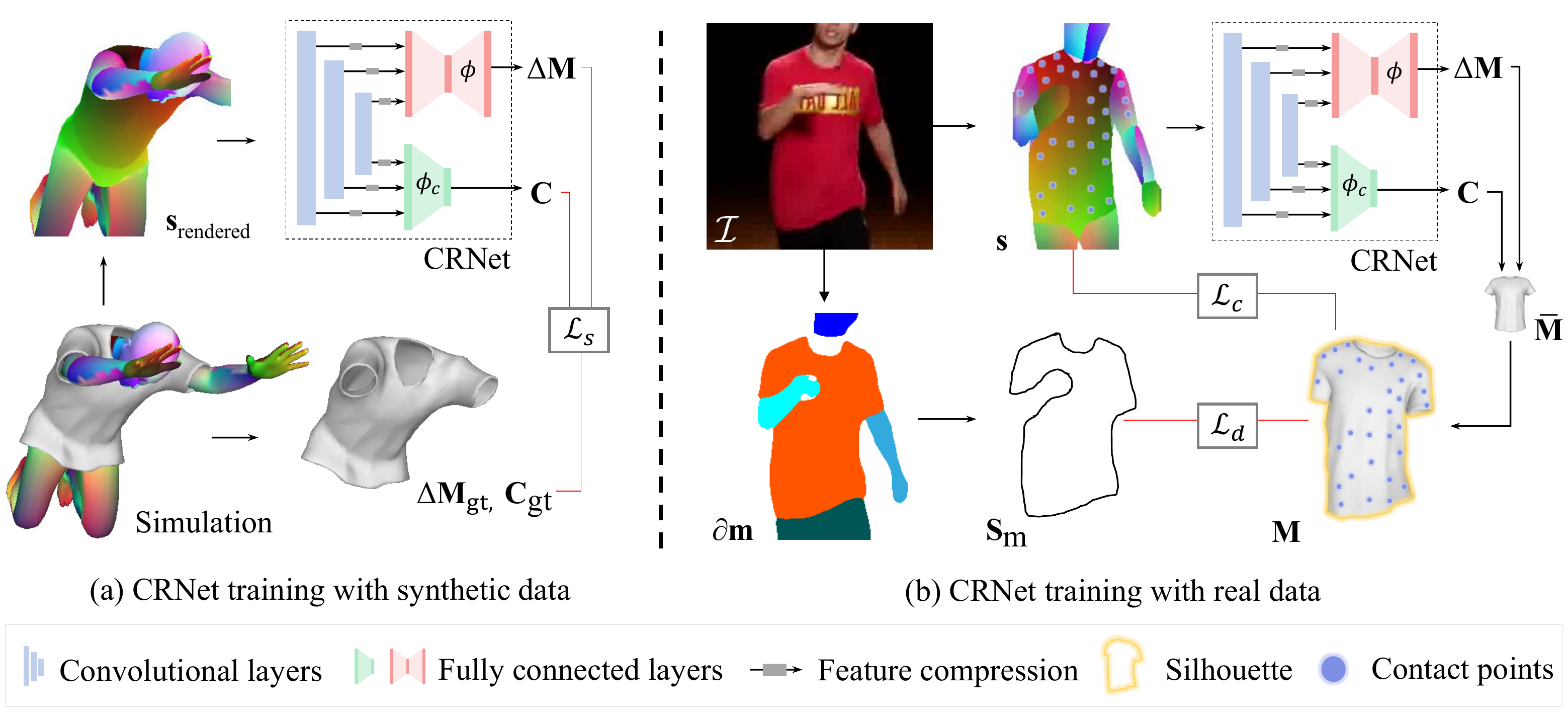}
	\end{center}	
	\vspace{-4mm}
	\caption{\small Given a densepose~$\mathbf{s}$ (synthesized or detected), CRNet extracts visual features (blue blocks) to regress a camera pose $\mathbf{C}$ and cloth deformation field $\Delta\mathbf{M}$ using $\phi$ (red blocks) and $\phi_C$ (green blocks), respectively. CRNet is fully-supervised with $\mathcal{L}_s$ using ground-truth synthetic data, and it adapts the weight of $\phi$ and $\phi_\mathbf{C}$ from synthetic to real-world data by minimizing the contact point correspondence distance $\mathcal{L}_c$ and semantic clothing boundary distance $\mathcal{L}_d$.}  
	\label{fig:training}
	\vspace{-.3cm}
\end{figure*}

We cast the problem of clothes retargeting as fitting the template clothing such that it can best explain the geometry of the clothing in the target image:
\begin{align}
    \mathbf{m} \leftrightarrow \Pi\left(\mathbf{M};\mathbf{R},\mathbf{t}\right), \label{Eq:retarget}
\end{align}
where $\Pi\left(\mathbf{M};\mathbf{R},\mathbf{t}\right)$ is the 2D projection of the $\mathbf{M}$ onto the affine camera with its spatial rotation $\mathbf{R}\in SO(3)$ and translation $\mathbf{t}\in \mathds{R}^3$. Note that $\mathbf{m}\in \mathds{R}^{M_m\times2}$ is the clothing in the target image which differs from the template model in shape, material, and topology. Therefore, $\mathbf{m}\neq \Pi(\mathbf{M})$ where $M_m$ is the number of vertices of the clothing in the target image.

In the following sections, we describe a new formulation of semi-supervised learning which reconstructs $\mathbf{M}$ from the pose and shape of a body using large-scale synthetic data, and utilizes the unlabeled real images to validate the physical plausibility of $\mathbf{M}$ by fitting it into the target image. To this end, we design a new network called Clothes Retargeting Network (CRNet) that integrates these two components to generate a plausible deformation of the template clothing. The overview of our clothes retargeting pipeline is described in Fig.~\ref{fig:overview}.

\subsection{Clothes Reconstruction By Synthesis}

Consider 2D body pose and shape $\mathbf{s}\in \mathds{R}^{M_s\times 2}$ in the target image. To reconstruct the 3D deformation of the template clothing $\mathbf{M}$, it requires two steps: (1) reconstructing the 3D body and camera from the 2D projected body, i.e., $\mathbf{S} = h(\mathbf{s})$ and $\mathbf{C} = h_C(\mathbf{s})$ where $\mathbf{C} = (\mathbf{R}, \mathbf{t}, k)$ is the camera parameter and $k$ is the scale of the 3D body, i.e., the inverse depth; (2) reconstructing 3D deformation of the template clothing from the reconstructed 3D body (Equation~(\ref{Eq:deform})). A main pitfall of this sequential reconstruction, $\mathbf{s}\rightarrow\mathbf{S}\rightarrow\mathbf{M}$, is that it involves reconstructing a nonlinear parametric body in 3D similar to body reconstruction approaches~\cite{bogo2016keep,xiang2018monocular}, which is highly prone to error propagation, i.e., mis-estimation in 3D body reconstruction leads erroneous clothing deformation. Furthermore, the two functions $h$ and $f$ need to be learned in isolation.

An interesting observation is that it is possible to bypass the intermediate representation of the 3D body $\mathbf{S}$, i.e., regressing the clothing  deformation $\Delta \mathbf{M}$ directly from the 2D body pose and shape $\mathbf{s}$:
\begin{align}
    \phi(\mathbf{s};\mathbf{w}) = f\circ h, \label{Eq:direct1}
\end{align}
where $\phi:\mathds{R}^{M_s\times2}\rightarrow \mathds{R}^{M\times3}$ is a map from 2D body shape to the 3D deformation of the clothing. The map is parametrized by weights $\mathbf{w}$. This is a significant departure from existing approaches~\cite{pons2017clothcap,danvevrek2017deepgarment} that requires the 3D reconstruction of the body as a prerequisite of clothes reconstruction process. Similarly, the camera pose can be regressed from $\mathbf{s}$:
\begin{align}
    \phi_C(\mathbf{s};\mathbf{w}_C) = h_C, \label{Eq:camera}
\end{align}
where $\phi_C$ recovers the camera pose and scale from 2D body shape, parametrized by $\mathbf{w}_C$.

To learn $\mathbf{w}$ and $\mathbf{w}_C$, we leverage large scale synthetic data generated by accurate physical simulation that animates the clothing  deformation in relation to the body pose and shape. The synthetic data are composed of a set of tuples ($\mathbf{s}, \Delta \mathbf{M}, \mathbf{C}$) by projecting them onto diverse camera viewpoints. In practice, we use Euler angle representation and $\mathbf{t}=(t_x,t_y,0)$ as an affine camera projection model is used. See Section~\ref{sec:synthetic} for more details of the synthetic data generation. Given the synthetic data, we learn $\phi$ and $\phi_C$ by minimizing the following loss:
\begin{align}
    \mathcal{L}_s(\mathbf{w}, \mathbf{w}_C) = \sum_{i=1}^{N_s} \|\phi(\mathbf{s}_i)-\Delta \mathbf{M}_i\|^2 + \|\phi_C(\mathbf{s}_i)-\mathbf{C}_i\|^2\nonumber
\end{align}
where $N_s$ is the number of synthetic data samples.

\subsection{Self-supervised Real Image Adaptation}
The mapping functions, $\phi$ and $\phi_C$, model the geometric relationship between the 2D body and a 3D template clothing. However, it still lacks the mapping to a target image:
\begin{align}
    (\mathbf{m}, \mathbf{s}) = g(\mathcal{I}), \label{Eq:decouple}
\end{align}
where $g(\mathcal{I})$ extracts the clothing and body from the image~$\mathcal{I}$. 


One may use a dense body pose detector~\cite{alp2018densepose} to extract $\mathbf{s}$ and feed into $\phi$ and $\phi_C$ as an input. This however shows inferior performance due to the gap between the synthetic data and real world clothing deformation. We bridge this gap by leveraging two self-supervisory signals, which allows $\phi$ and $\phi_C$ to adapt to the real scenes.\\


\noindent\textbf{Contact Points} While the contact regions between the body and clothing can vary in general, there are few persistent contact surfaces that maintain the overall clothing shape. The distance between the body and clothing points on these contact surfaces could be used to constrain our problem. 
We manually specify such points (see implementation details in Section~\ref{sec:exp}) which allow us to measure the geometric error between the reconstruction and image: 
\begin{align}
    \mathcal{L}_c(\mathbf{w}, \mathbf{w}_C) = \sum_{i=1}^{N_r}\sum_{j=1}^{M}\delta_j \| \Pi (\mathbf{M}^j_i;\mathbf{R}_i,\mathbf{t}_i) - \mathbf{s}^{z(j)}_i\|^2
\label{contact_loss}
\end{align}
where $\mathbf{M}_i^j \in \mathds{R}^3$ is the $j^{\rm th}$ point of the deformed clothing computed by the $i^{\rm th}$ input image $\mathbf{M}_i = \phi(g(\mathcal{I}_i))$, and $\mathbf{s}^{z(j)}_i \in \mathds{R}^3$ is the body point corresponding to the $j^{\rm th}$ clothing point, i.e., $z(j)$ maps contact points between body and clothing. $\delta_j$ is Kronecker delta that produces one if the $j^{\rm th}$ clothing point belongs to the set of contact points, and zero otherwise. $N_r$ is the number of the real image data samples. Note that the camera parameter $\mathbf{R}$ and $\mathbf{t}$ is computed by $\phi_C(g(\mathcal{I}_i);\mathbf{w}_C)$.\\

\noindent\textbf{Clothing Silhouette} Although there is no explicit correspondence between the template clothing and the clothing of a person in the input image, it is reasonable to assume that their silhouette of the template clothing must match to that of the target clothing as they undergo similar dynamics~\cite{habermann2020deepcap,habermann2019livecap,xu2018monoperfcap}. We formulate this as a Chamfer distance transform between two sets of silhouettes:


\begin{align}
    D(\mathcal{S}_{\mathbf{m}}, \mathcal{S}_{\mathbf{M}}) = \sum_{\mathbf{x}\in \mathcal{S}_{\mathbf{m}}} \underset{\mathbf{y} \in \mathcal{S}_{\mathbf{M}}}{\operatorname{min}}~\|\mathbf{x}-\mathbf{y}\|^2 + 
    \sum_{\mathbf{y} \in \mathcal{S}_{\mathbf{M}}} \underset{\mathbf{x}\in \mathcal{S}_{\mathbf{m}}}{\operatorname{min}}~\|\mathbf{x}-\mathbf{y}\|^2\nonumber
\end{align}
where $\mathcal{S}_{\mathbf{m}}$ is the set of silhouette points on the presented clothing, and $\mathcal{S}_{\mathbf{M}}$ is the set of silhouette points on the template clothing. We use this Chamfer distance measure to align the template clothing to the target clothing by minimizing:
\begin{align}
    \mathcal{L}_d (\mathbf{w}, \mathbf{w}_C) = \sum_{i=1}^{N_r} D(\mathcal{S}_{\mathbf{m}_i}, \mathcal{S}_{\mathbf{M}_i}).
\label{boundary_loss}
\end{align}
In practice, we remove the correspondences whose Chamfer distance is larger than the predefined threshold to deal with the mismatch of the clothing types between the source and target clothing. For example, in Fig~\ref{fig:training}(b), the boundary points on the right arm are removed
as outliers because their Chamfer distance with 3D clothing boundary (visualized with yellow) is larger than a threshold.

\subsection{CRNet Design}
\label{sec:crnet}
We design a new network called Clothes Retargeting Network (CRNet) that generates plausible deformations of a template clothing from an image as shown in Fig.~\ref{fig:training}. CRNet is trained to learn $\mathbf{w}$ and $\mathbf{w}_C$ in $\phi$ and $\phi_C$, respectively, by minimizing three losses ($\mathcal{L}_s$, $\mathcal{L}_c$, and $\mathcal{L}_d$). 
For synthetic data, the 3D clothing deformation $\Delta \mathbf{M}$ and camera pose $\mathbf{C}$ are directly regressed from 2D synthetic body pose and shape $\mathbf{s}$, where $\Delta \mathbf{M}$ and $\mathbf{C}$ are fully supervised with synthetic ground-truth data with $\mathcal{L}_s$. For real images, the 2D body pose and shape $\mathbf{s}$ is replaced by those detected from real-world images using a dense pose detector~\cite{alp2018densepose}, and $\mathbf{M}=\overline{\mathbf{M}}+\Delta \mathbf{M}$ is projected onto the image using the camera parameter $\mathbf{C}$ to measure $\mathcal{L}_c$ and $\mathcal{L}_d$. We design $\mathbf{w}$ and $\mathbf{w}_C$ to share the image feature through skip connections, allowing $\phi$ and $\phi_C$ to access both local and global body shape in the image. The network details can be found in the supplementary material.

\subsection{Online Refinement}

For a testing image, we further refine the weights of decoders $\mathbf{w}$ and $\mathbf{w}_C$ by making use of the self-supervised loss $\mathcal{L}_c, \mathcal{L}_d$. In practice, we hierarchically update the weights of CRNet as follows: (1) we update the weight for camera pose $\mathbf{w}_C$ using the contact correspondence loss $\mathcal{L}_c$ while $\mathbf{w}$ is fixed. (2) We update $\mathbf{w}$ using $\mathcal{L}_c$ to find the best non-rigid clothing deformation given a 2D body pose while $\mathbf{w}_C$ is fixed. (3) We update both $\mathbf{w}_C$ and $\mathbf{w}$ using $\mathcal{L}_c$ and $\mathcal{L}_d$ such that CRNet learns to predict the best clothes retargeting by incorporating the semantic boundary as well as body-to-clothing contact points. The detailed algorithm flow of the online refinement step with specific parameter settings is summarized in the supplementary material.


\vspace{5mm}
\subsection{Synthetic Data Generation}\label{sec:synthetic}
We create a large collection of synthetic data using physics-based simulation. Given a 3D clothing mesh model $\overline{\mathbf{M}}$ with T-pose $\overline{\mathbf{S}}$, we deform the clothing by changing pose and shape of clothing using contact dynamics~\cite{li2018implicit}. We establish the contact correspondences between body and clothing based on T-pose and deform the clothing by minimizing following measure:

\begin{align}
E(\Delta{\mathbf{M}};\mathbf{S})=E_{c}+\lambda_{r}E_{r}+\lambda_{s}E_{s}, 
\end{align}
where $E_{c}$ is the contact constraint that encourages the clothing to move along with the body through the contact points similar to Equation~(\ref{contact_loss}):        
\begin{align}
E_{c}(\mathbf{M}, \mathbf{S}) = \sum_{j=1}^{M}\delta_j \| \mathbf{M}^j - \mathbf{S}^{z(j)}\|^2.
\end{align}
$E_{r}$ and $E_{s}$ are the regularization terms, and $\lambda_{r}, \lambda_{s}$ control the importance of each term. $E_{r}$ prevents from collapsing the clothing structure by enforcing rigidity at each vertex, i.e, $E_{r}(\mathbf{M}) = \sum^{\textit{M}}_{i}\sum_{j\in \mathcal{N}(i)}\|({\mathbf{M}}_i-{\mathbf{M}}_j) - \mathbf{R_i}(\overline{\mathbf{M}}_i-\overline{\mathbf{M}}_j)\|^2$ where $j$ is the index of neighboring vertices, and $E_{s}$ ensures the deformation that minimizes Laplacian residual~\cite{sorkine2007rigid}, i.e., $E_{s}(\mathbf{M}) = \|L(\overline{\mathbf{M}}) - L(\mathbf{M})\|^2$ where $L(\mathbf{M})$ is a Laplacian of $\mathbf{M}$.

\begin{figure}[t]
	\begin{center}
		\includegraphics[width=3.3in]{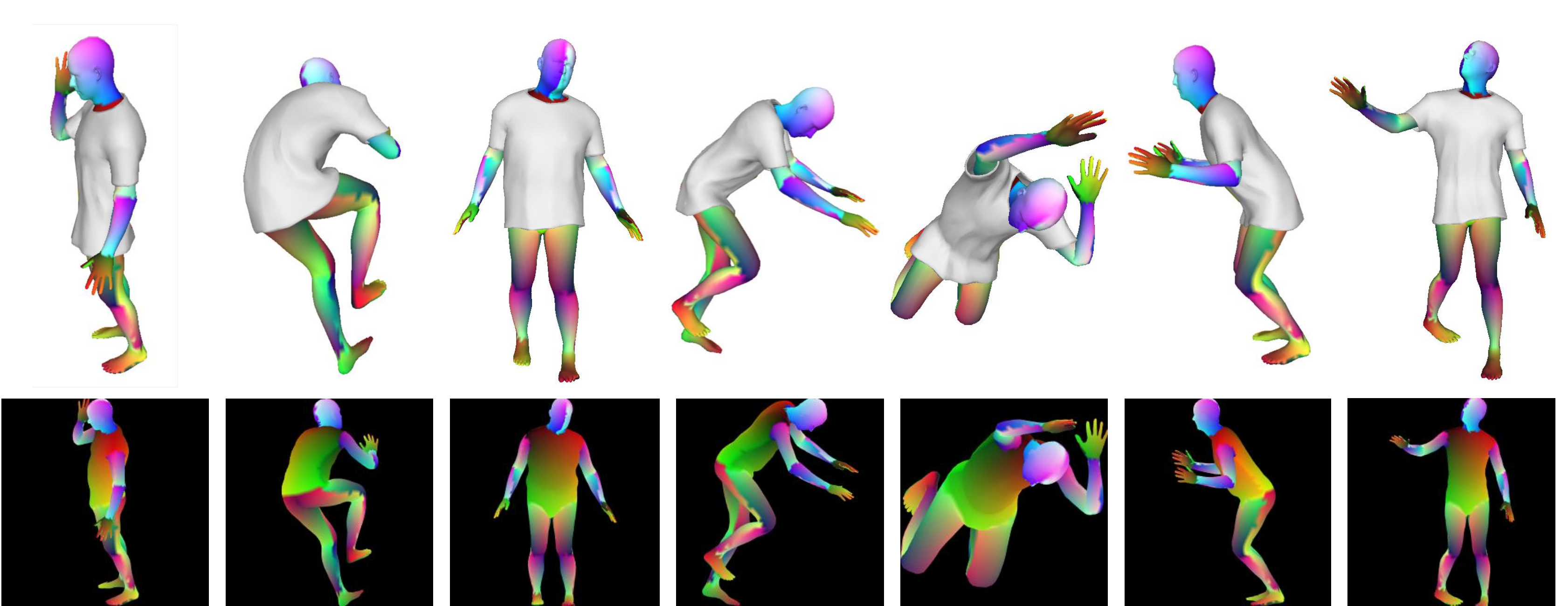}
	\end{center}	
	\vspace{-.5cm}
	\caption{\small Examples of the synthetic data generation process.  (Top) Our synthetic 3D clothing controlled by SMPL~\cite{loper2015smpl} body model, and (Bottom) synthesized 2D dense body pose image generated by 2D projection of the color-coded SMPL body model. 
	}
	\vspace{-3mm}
	\label{fig:synthetic}
\end{figure}

To control the body model for generating $\mathbf{S}$, we use SMPL body model~\cite{loper2015smpl} with the natural human body pose and shape parameters~\cite{varol2017learning}. The rigid body pose and shape are randomly created as well as camera pose~$\mathbf{C}$ as shown in Fig.~\ref{fig:synthetic}. To generate synthetic 2D dense pose map $\mathbf{s}$, we project the color-coded SMPL model with the visibility using the Z-buffer algorithm~\cite{wand2001randomized}. 
Fig.~\ref{fig:synthetic} describes the examples of synthetic 3D clothing deformation and 2D dense body pose map.

\section{Experiments}\label{sec:exp}

For the qualitative evaluation, we demonstrate the results from various examples with arbitrary poses using different clothing templates. For the quantitative evaluation, we compare our single view retargeting results with multiview reconstruction results. We also provide ablation studies of our online refinement modules.

\begin{figure}[t]
	\begin{center}
		\includegraphics[width=3.15in]{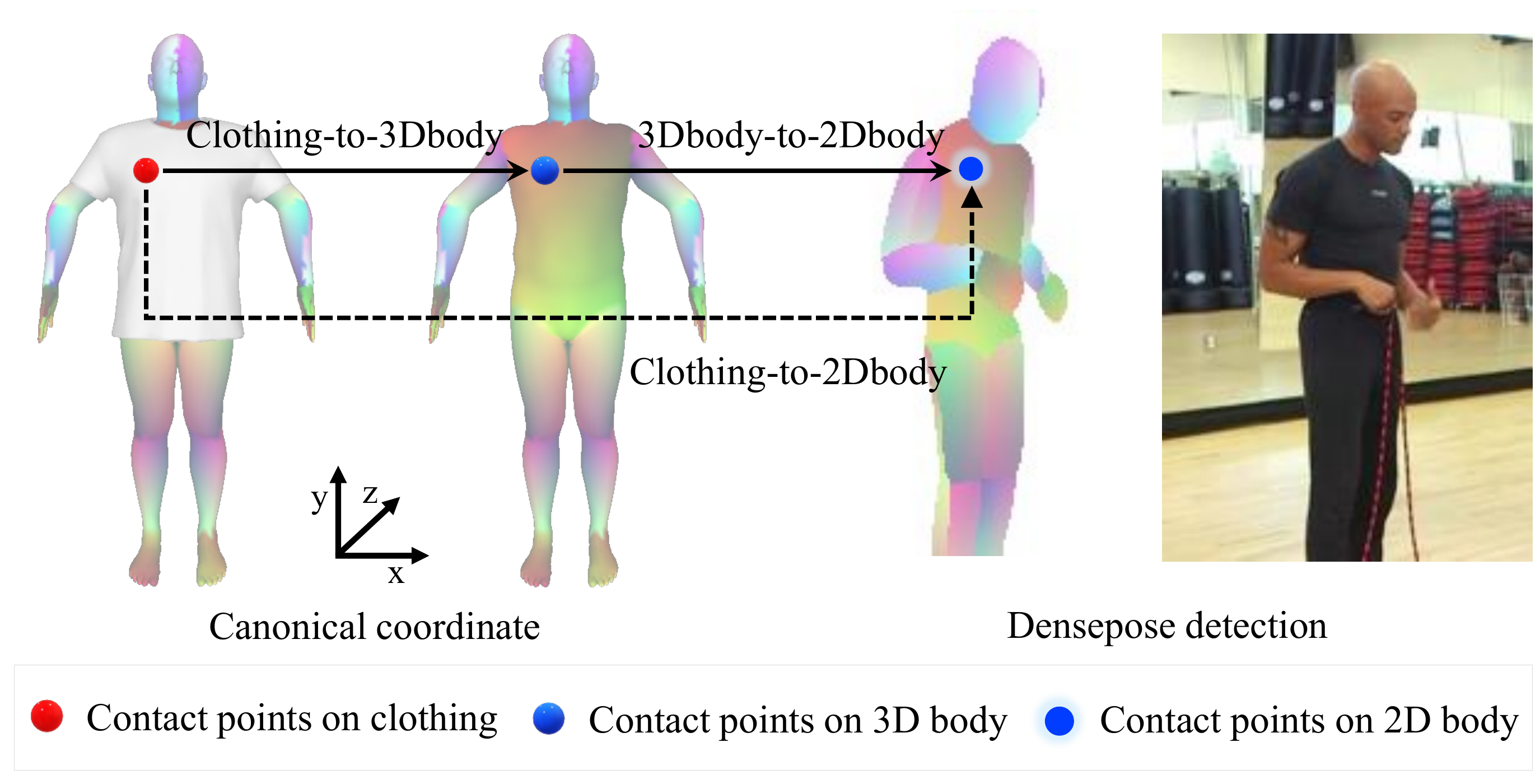}
	\end{center}	
	\vspace{-0.5cm}
	\caption{\small The process to pre-define the contact points between 3D clothes and 2D densepose for contact points loss.}
	\label{Fig:contactpoint}
		\vspace{-3mm}
\end{figure}

\noindent \textbf{Implementation Details}
Given a clothing template $\overline{\mathbf{M}}$ and the initially aligned body model $\overline{\mathbf{S}}$, we first generate a large-scale synthetic training dataset as described in Section~\ref{sec:synthetic}, where 100,000 pairs of ($\mathbf{s}$, $\Delta\mathbf{M}$, $\mathbf{C})$ are enough to fully supervise the network. We further prepare around 5,000 real-world human body images from~\cite{andriluka14cvpr}, and random augmentation (rotation, translation, scale variation, flip) is applied at every training iteration. During training, we alternatively train the real and synthetic data, where CRNet is trained with Stochastic Gradient Decent (SGD). To obtain the clothing boundary for silhouette loss $\mathcal{L}_d$, we compute the clothing silhouette using semantic segmentation~\cite{lin:2017} that identifies body/clothing parts as shown in Fig.~\ref{fig:training},


The contact point correspondences between the clothing template and 2D densepose are manually specified as follows: 1) In the canoncial 3D space, where the clothes and body model are aligned, we identify the ``clothing to 3D body'' contact points based on the closest distance between two vertices sets. If the distance is smaller than the pre-defined threshold, the vertex is considered as contact point. 2) We define the ``clothing to 2D body'' contact points by bypassing the intermediate ``3D to 2D body'' correspondences~\cite{alp2018densepose} as described in Fig.~\ref{Fig:contactpoint}.

\begin{figure}[t]
	\vspace{-5mm}
	\begin{center}
\includegraphics[width=0.48\textwidth]{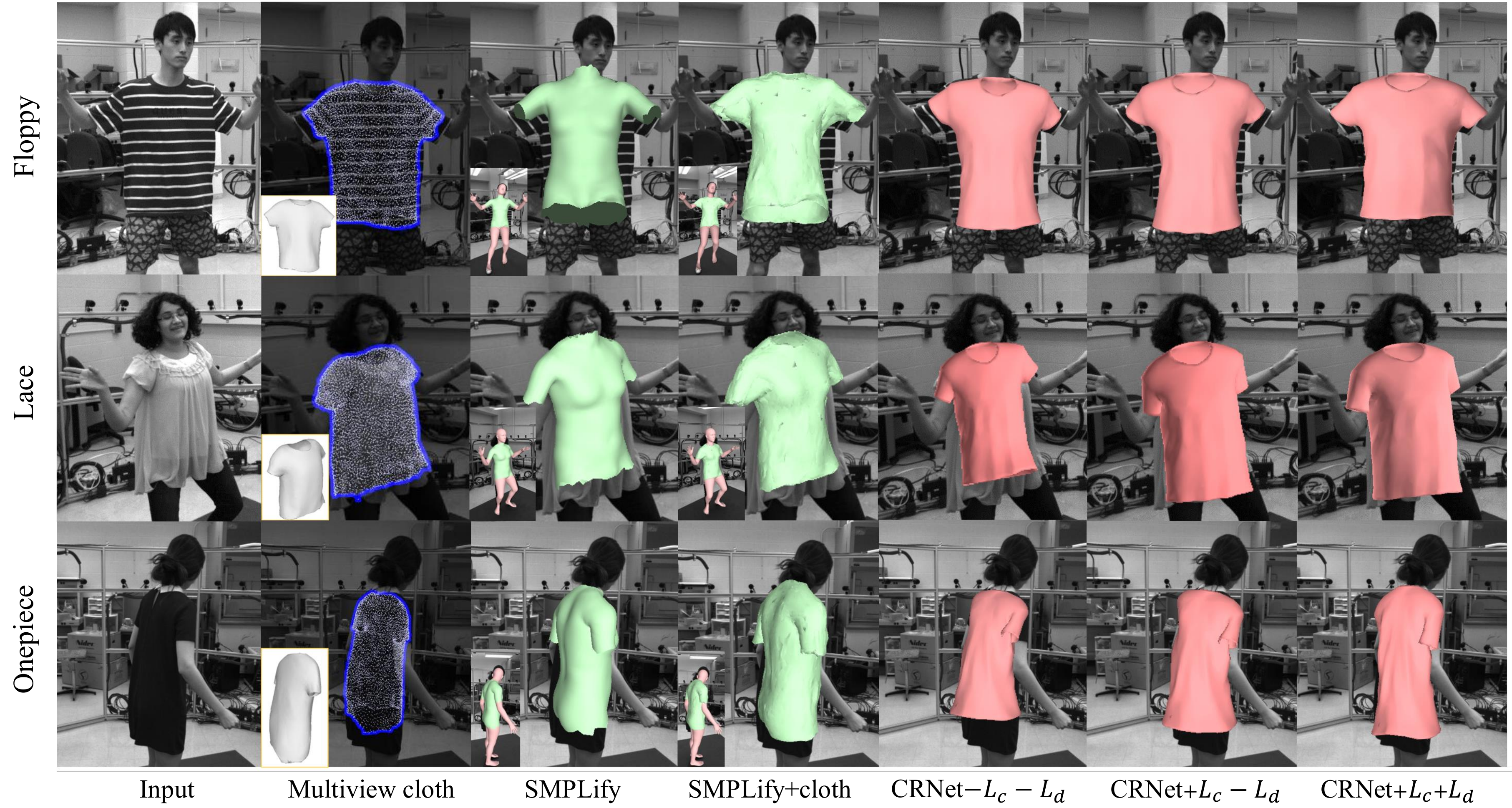}
	\end{center}	
	\vspace{-7mm}
	\caption{\small Qualitative comparison matched to the Table~\ref{table:exp}.}
	\label{fig:qual_res}
\end{figure}

\noindent \textbf{Evaluation Dataset} 
To quantitatively evaluate the performance of CRNet, we use the public multiview clothing dataset, HUMBI~\cite{yu2020humbi}, as a proxy of ground-truth, which provides 3D clothing fitting results on a hundred of people and the associated multiview images. Among them, we choose seven sequences where the same clothing model as ours is used for these sequences. In Table~\ref{table:exp}, the subject number denotes the sequence ID on this benchmark. For our analysis, we capture three more new and challenging sequences from a large-scale multiview setting~\cite{shin20183d} as shown in Fig.~\ref{fig:qual_res}. \textbf{Lace}: Women with lacy shirts dance intensely from side to side. In this scenario, the freedom of cloth deformation is higher than with normal T-shirts. \textbf{One piece}: A woman with a short one-piece performs her movements. Here, the length of the clothing is longer than our clothing model. \textbf{T-shirt}: This sequence contains a man who wears a big T-shirt, requiring bigger secondary deformation. To fit the clothing template model to the multiview 3D reconstruction, we use the same algorithm as the state-of-the-art clothing capture method~\cite{pons2017clothcap}, where the manual T-pose prior for clothing area segmentation is replaced with automatic 3D clothing segmentation using a 3D semantic map~\cite{shin20183d}. The details of the method to capture the 3D clothing are described in the supplementary materials.

\noindent \textbf{Evaluation Metric} 
The different camera model assumptions between the calibrated multiview system, with perspective projection model, and a single view weak-perspective model, with orthogonal projection, produces a coordinate and scale ambiguity between them. For this reason, we evaluate the accuracy of the retargeted 3D clothing in the reprojected 2D image space based on the Euclidean distance as follows:
\begin{align}
 \sum_{i=1}^M \delta_i^v \|\widetilde{\mathbf{m}}_i - \Pi(\mathbf{M}_i;\mathbf{R},\mathbf{t})\|^2,
\label{eq:metric1}
\end{align}
where $\widetilde{\mathbf{m}}_i \in \mathds{R}^2$ is the ground truth of the $i^{\rm th}$ 2D clothing point, and $\delta_i^v$ is Kronecker delta that produces one if $\mathbf{M}_i$ is visible to the camera, and zero otherwise. The visibility is accurately computed using the Z-buffer algorithm~\cite{wand2001randomized}. 

Note that $\widetilde{\mathbf{m}}$ is 2D projection of the clothing template, and therefore, the clothing index is consistent with $\mathbf{M}$. The ground truth is obtained by reconstructing the clothing in 3D using multiple cameras and projecting onto the testing camera. To prevent the inconsistent pixel error scale (e.g., the pixel distance error proportionally increases with respect to the scale of the object in the image), we evaluate the error with respect to the network input image scale, where the input is made by cropping the original image around the clothing area with the same aspect ratio, i.e., 1:1 ratio, and then resizing into 256. We then finally normalize the error with the size of the image. In this way, we can evaluate the errors with consistent pixel-scale space.

{
\renewcommand{\tabcolsep}{4pt} 

\begin{table}[t]
\centering

\scriptsize
\begin{tabular}{l|c|c|c|c|c}
\hline

 & SMPLify &  SMPLify \cite{bogo2016keep} & \textbf{CRNet} & \textbf{CRNet} & \textbf{CRNet} \\
 & \cite{bogo2016keep} & +clothing & \textbf{-$\mathcal{L}_c$-$\mathcal{L}_d$} & \textbf{+$\mathcal{L}_c$-$\mathcal{L}_d$} &\textbf{+$\mathcal{L}_c$+$\mathcal{L}_d$}
\\

\hline
Subj.83& {4.76}$\pm$1.54 &4.68$\pm$1.51 &{5.4}$\pm$2.5&{4.15}$\pm$1.84 & \textbf{{4.11}}\textbf{$\pm$1.82 }\\
\hline
Subj.89& {4.75}$\pm$1.18 & 4.74$\pm$1.18& {5.61}$\pm$2.66  & {4.81}$\pm$1.41 & \textbf{{4.3}\textbf{$\pm$1.27}}\\
\hline
Subj.123& {5.97}$\pm$1.09&5.95$\pm$1.12& {7.16}$\pm$2.55 &   {5.30}$\pm$1.27  & \textbf{{5.15}}\textbf{$\pm$1.23} \\
\hline
Subj.130& {4.33}$\pm$1.79  &4.36$\pm$1.78 & {5.64}$\pm$3.02 & {4.31}$\pm$1.60  & \textbf{{4.27}}\textbf{$\pm$1.55} \\
\hline
Subj.133& {6.12}$\pm$2.09 &6.17$\pm$2.09  & {7.82}$\pm$2.74 & \textbf{{5.57}} \textbf{$\pm$1.25}  & {{5.69}}{$\pm$1.18} \\
\hline
Subj.134& {4.54}$\pm$0.97 &4.60$\pm$0.95 &{6.25}$\pm$2.81 & {5.08}$\pm$1.58  & \textbf{{3.38}}\textbf{$\pm$1.2} \\
\hline
Subj.138& {4.32}$\pm$1.51&4.26$\pm$1.52 & {5.12}$\pm$2.39 & {4.83}$\pm$0.93  & \textbf{{4.23}}\textbf{$\pm$1.25} \\
\hline
\hline
Lace & 6.38$\pm$1.77  &6.42$\pm$1.75 & 7.74$\pm$3.46 & {6.61}$\pm${1.59}  & \textbf{{5.31}$\pm$\textbf{{1.08} }}\\
\hline
Onepiece  & 7.54$\pm$1.04&7.57$\pm$1.04 & {8.3}$\pm$2.3 & {8.01}$\pm$2.19  & \textbf{{6.88}}\textbf{$\pm$1.32}\\
\hline
T-shirt & 6.19$\pm$1.137 & 6.10$\pm$1.14 & {6.97}$\pm$1.27 & {6.05}$\pm$0.79  & \textbf{{5.30}}\textbf{$\pm$0.91}\\
\hline
\hline
Average & 5.50$\pm$1.41 &5.49$\pm$1.42 & {6.60}$\pm2.57$& {5.47}$\pm1.44$   & \textbf{{4.86}}\textbf{$\pm$1.28}\\
\hline
\end{tabular}
\vspace{-2mm}
\caption{\small Quantitative evaluation on the multiview 3D clothing data. The average score and its standard deviation are reported with respect to the entire sequence. (Unit: \%)}
\label{table:exp}
\end{table}
}

\noindent \textbf{Baseline} We conduct ablation study of our self-supervised losses and compare with the state-of-the-art body pose and shape reconstruction methods. For the ablation study, three models are validated to investigate the importance of each loss: \textbf{CRNet-$\mathcal{L}_c$-$\mathcal{L}_d$}: CRNet trained only with synthetic data without any self-supervised loss. \textbf{CRNet+$\mathcal{L}_c$-$\mathcal{L}_d$}: CRNet trained with synthetic and real data including contact point correspondence loss. \textbf{CRNet+$\mathcal{L}_c$+$\mathcal{L}_d$} (Ours): CRNet trained with synthetic and real data including both contact point correspondence loss and clothing boundary loss. For comparative evaluation, 
we compare our method with the approaches of body model fitting to a single view image (SMPLify~\cite{bogo2016keep}) and afterwards fitting the clothing model by using Laplacian deformation~\cite{sorkine2007rigid} given prescribed body contact points (SMPLify~\cite{bogo2016keep}+clothing). To handle the topological difference in clothing and body model, e.g., the number of vertices and index orders are different, we manually define the correspondence between two models in the canonical space based on the closest vertex distance as visualized with green color in Fig.~\ref{fig:qual_res}-\cite{bogo2016keep}. The overall description of this evaluation is summarized in Table~\ref{table:exp} and Fig.~\ref{fig:qual_res}.

\begin{figure}[t]
		\vspace{-2mm}
	\begin{center}f
		\hspace{0mm}
\includegraphics[width=0.45\textwidth]{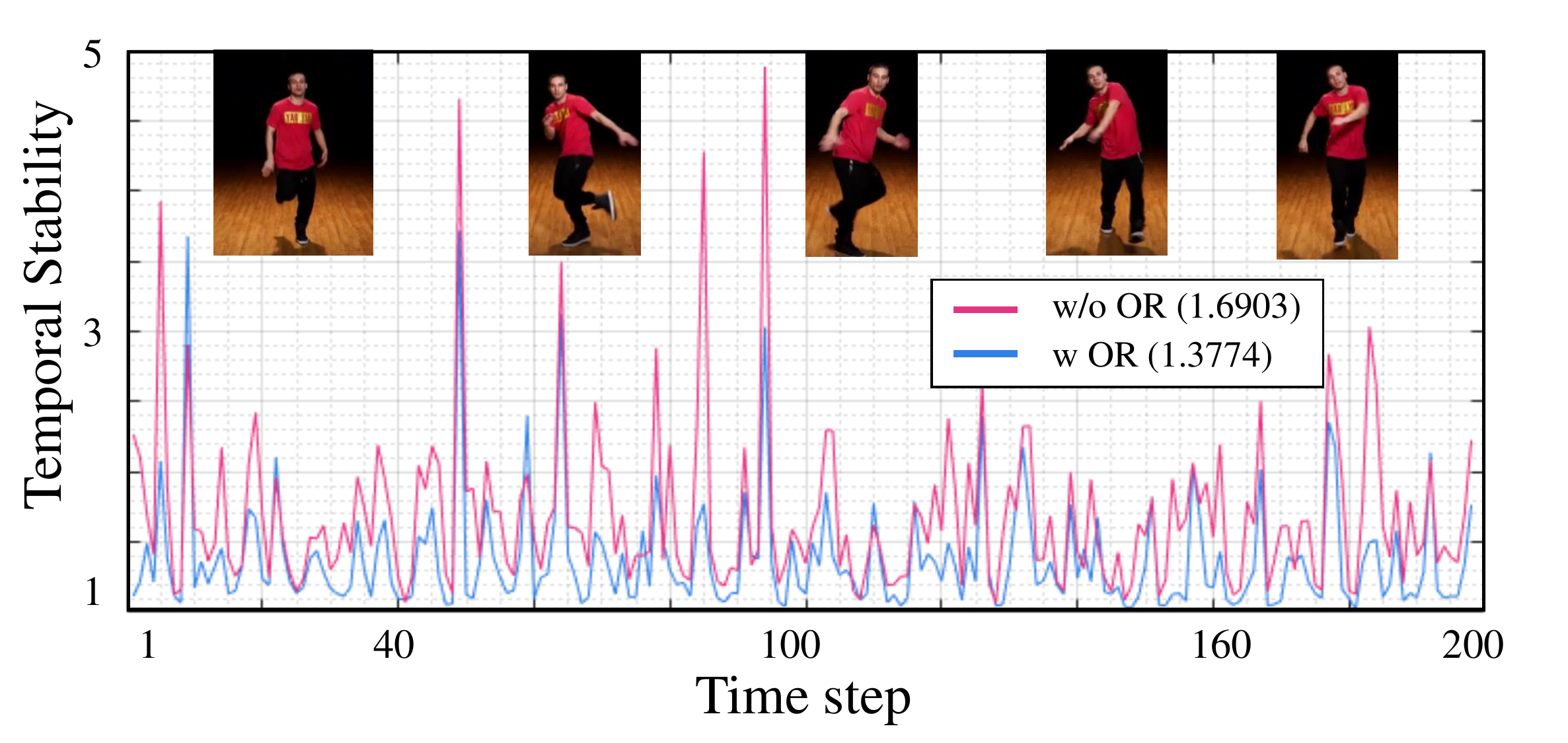}
	\end{center}	
	\vspace{-8mm}
	\caption{\small Ablation studies on the effect of with (w) and without (w/o) our online refinement (OR) based on the temporal stability metric. The average score with respect to the entire evaluation dataset is reported. Note that neither temporal smoothing nor tracking modules are applied. }
	\label{fig:temporal-consist}
	\vspace{-4mm}
\end{figure}

	\vspace{-2mm}
\subsection{Comparative Analysis}
	\vspace{-1mm}
As we summarized in Table~\ref{table:exp}, our method performs the best with respect to the average score. Based on the comparison of \textbf{CRNet-$\mathcal{L}_c$-$\mathcal{L}_d$}  with \textbf{CRNet+$\mathcal{L}_c$-$\mathcal{L}_d$}, we notice that training only with synthetic data is not fully generalizable to the real-world data, whereas the contact point loss $\mathcal{L}_c$ plays a key role to adapt the weight of CRNet to real data. The comparison between \textbf{CRNet+$\mathcal{L}_c$-$\mathcal{L}_d$} and \textbf{CRNet+$\mathcal{L}_c$+$\mathcal{L}_d$} shows that the semantic boundary helps the CRNet to capture the details on the clothing.
In Table~\ref{table:exp}, the keypoint detection error leads to inaccurate SMPL prediction (SMPLify~\cite{bogo2016keep}), and then clothing shape recovery (SMPLify~\cite{bogo2016keep}+clothing) sequentially. 
In contrast, our self-supervision approach refines the clothing to best match to the target image which significantly mitigates the prediction error, which highlights the performance of our clothes retargeting network.\\




\begin{figure*}[t]
	\begin{center}
		\includegraphics[width=0.9\linewidth]{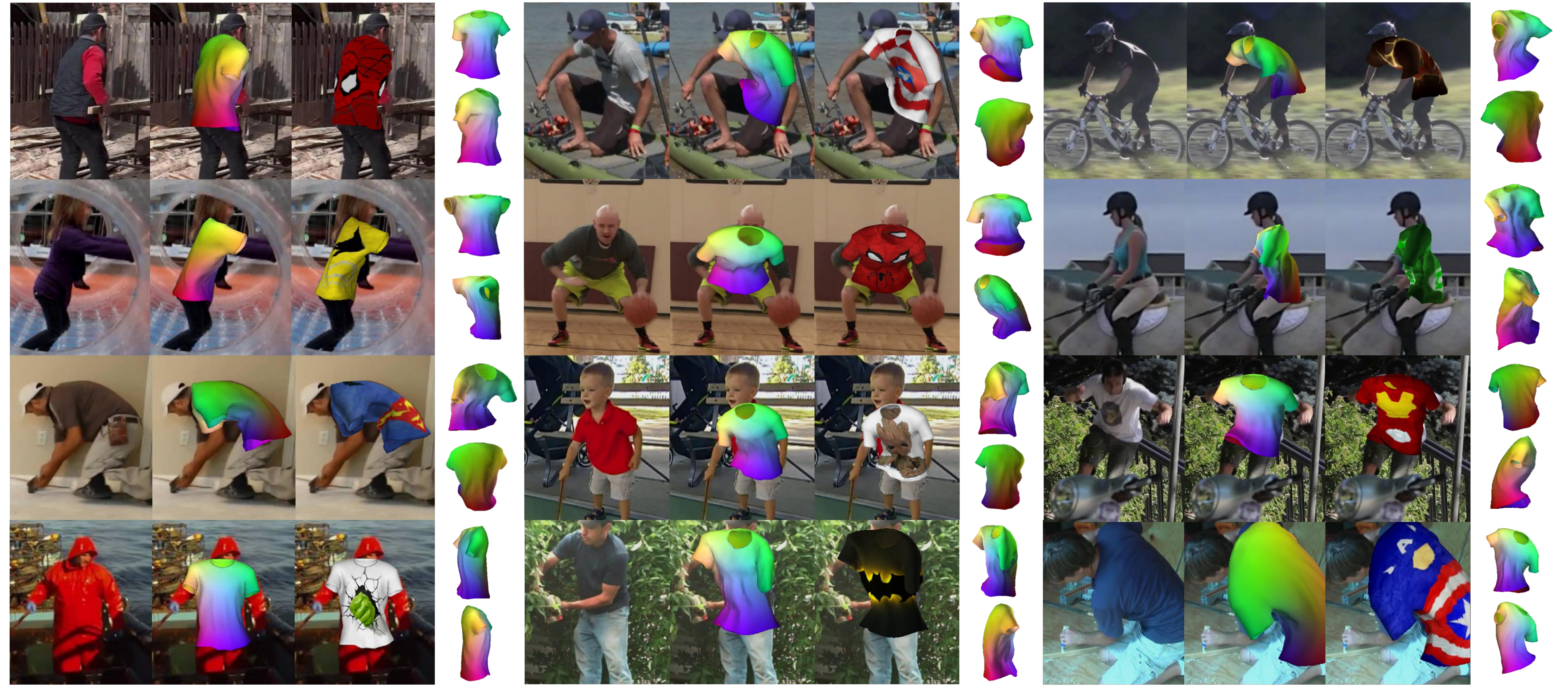}
	\end{center}	
	\vspace{-6mm}
	\caption{\small Results of clothes retargeting to a person with different poses and appearances. The retargeted clothing is visualized with the texture coordinates and the newly mapped textures. For each subject, the last column shows the retargeted clothing seen from different angles.
	}
	\label{fig:qualitative}
	\vspace{-.3cm}
\end{figure*}

\begin{figure*}[t]
	\begin{center}
		\includegraphics[width=0.9\linewidth]{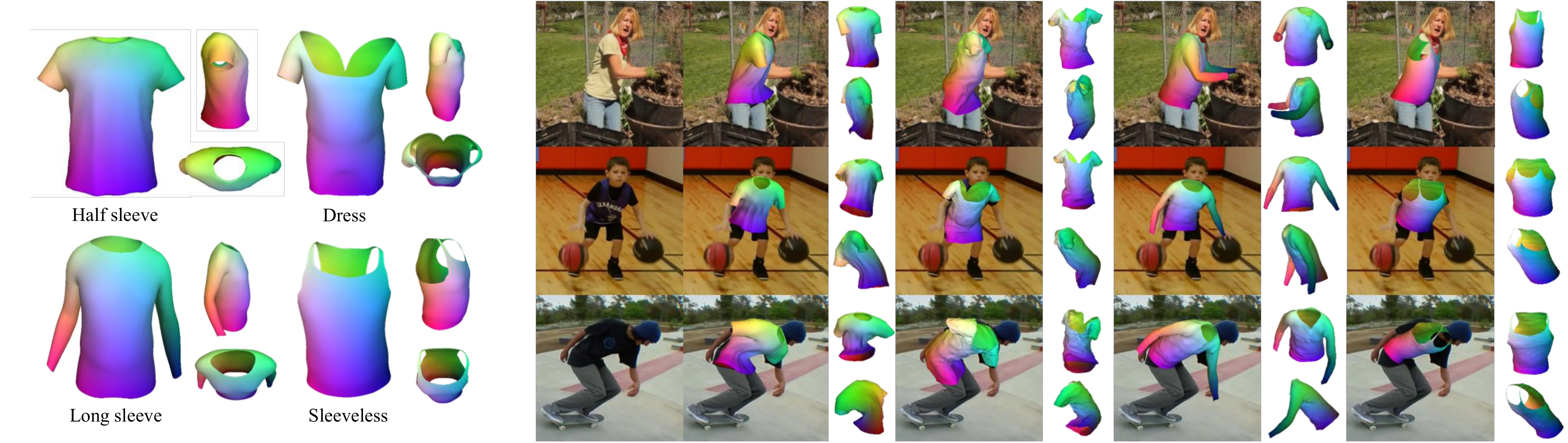}
	\end{center}	
	\vspace{-6mm}
	\caption{\small Clothes retargeting with various clothing templates. CRNet can learn the deformation of the various clothing templates with the automatically synthesized unlimited number of dataset and then adapt to real images with self-supervision. For each example, the last column shows the retargeted clothing seen from different angles.
	}
	\label{fig:qualitative_template}
	\vspace{-.3cm}
\end{figure*}

\noindent \textbf{Effect of Online Refinement} We observe that if the domain of CRNet is refined well on the online testing image, the retargeted clothing output has better visual realism spatially, i.e., the clothing is visually aligned well in a single image, which is validated in Table~\ref{table:exp}, as well as temporally, i.e., the clothing has smooth motion for sequential images without utilizing a temporal smoothing or tracking module. Based on this observation, we perform the additional ablation test on our online refinement algorithm with the metric of temporal stability as similar to~\cite{Yoon_2019_CVPR} defined as follows:
\begin{align}
\sum_{i=1}^{M}\frac{\left\|\mathbf{M}^{t+1}_{i}-\mathbf{M}^{t}_{i}\right\|_{2}+\left\|\mathbf{M}^{t}_{i}-\mathbf{M}^{t-1}_{i}\right\|_{2}}{\left\|\mathbf{M}^{t+1}_{i}-\mathbf{M}^{t-1}_{i}\right\|_{2}},
\label{eq:temporal_stability}
\end{align} 
where $t$ denotes the time step. Because the Equation~(\ref{eq:temporal_stability}) is a relative measurement that does not require the ground-truth, we evaluate our CRNet on video clips chosen from Youtube. As described in Fig.~\ref{fig:temporal-consist}, our online refinement step can enforce temporal stability, where the average score is 1.37, which is close to the lower bound of our metric (=1).

\subsection{Qualitative Results}
We demonstrate the performance of our clothes retargeting method by showing extensive results on in-the-wild images of people in clothes. Fig.~\ref{fig:qualitative} illustrates the recovered clothing models that are physically and visually plausible in the presence of a large variation of shapes, deformations, and viewpoints. We can also change the clothing appearance by mapping a random texture onto the retargeted clothing model due to the nature of the fixed model topology as described in Fig.~\ref{fig:qualitative}. More results from a variety of examples will be shown in supplementary material. \\ 
\textbf{Application to a New Clothing Model} \ \
While a T-shirt model is used for a proof-of-concept, our method is agnostic to the clothing shape. For example, a new onepiece dress model as shown in Fig.~\ref{fig:qualitative_template} can be automatically overlaid onto a person in an image with a few minor modifications of our clothes retargeting pipeline: 1) The output dimension of CRNet should be matched to the number of vertices on a new model. 2) The contact point correspondences between the body model and a new clothing model are re-prescribed in the canonical space. 3) The class of semantic boundary should reflect the type of clothing model, e.g., cloth top and dress classes are used for the onepiece model. Fig.~\ref{fig:qualitative_template}  shows the results with various  models.

	




\section{Conclusion}
We present a learning based approach to retarget a given 3D clothing mesh onto a person in a single 2D image. To accurately transfer and fit the clothing template into the image, we estimate the most plausible deformation of the clothing model that best aligns the pose of the body with the semantic boundary of the clothing silhouette in the image, based on semi-supervised training. We show that our method can provide an accurate retargeting solution that is comparable to multi-view based approaches, and we demonstrate how the method can handle arbitrary images. 

It is also worth to note a few of limitations in our method. First, while we predict the entire clothing from a partially observable clothing area, occlusion is still challenging similar to other single view based reconstruction solutions. Second, the network is specific to each shape and therefore, the network should be trained in conjunction with synthetic data generation. 



{\small
\bibliographystyle{ieee}
\bibliography{arxiv.bbl}
}

\clearpage

\setcounter{section}{0}
\def\thesection{\Alph{section}}
\renewcommand{\thesubsection}{\thesection.\arabic{subsection}}

\twocolumn[{%
\maketitle
	\begin{center}
		\centering
	\includegraphics[width=6.2 in]{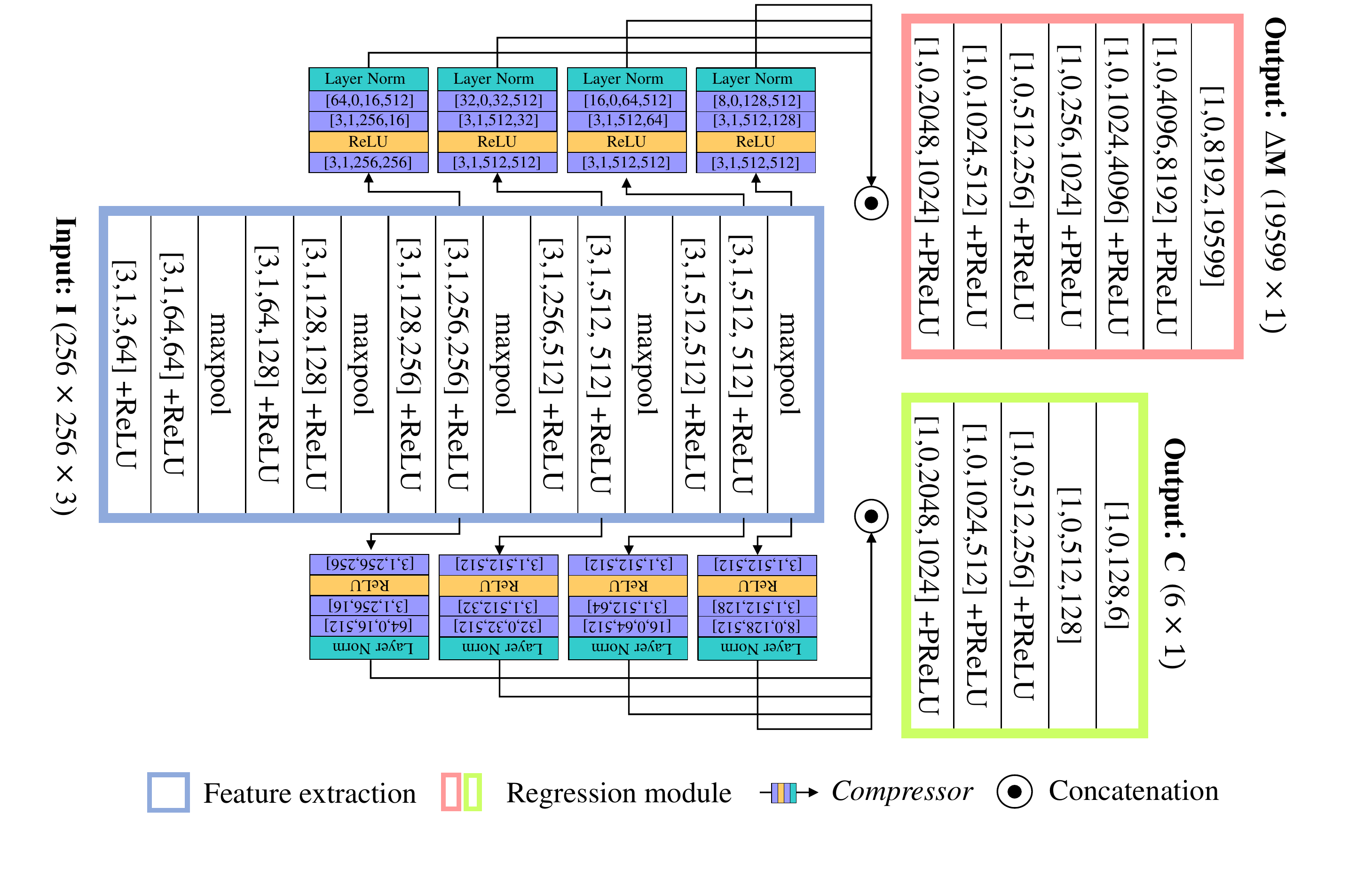}

	\captionof{figure}{The implementation details of our Clothes Retargeting Network (CRNet). In the convolutional block, the filter property is defined in the following order: [filter size, zero-padding size, input channel, output channel]. }
\label{fig:supple_network}
\end{center}	
}]

\section{Overview}

This supplementary material provides additional details of the proposed method, more discussions and qualitative results to support the main paper.

In Section~\ref{sec:crn}, we provide the implementation details of the CRNet. In Section~\ref{sec:algoritm}, we introduce the algorithm details of the online refinement (corresponding to the Section~3.4 in the main paper) with specific parameter setting. In Section~\ref{sec:multiview}, we also show the extra details of the process to capture our multi-view data used for the evaluation in Section.~4 of the main paper. In Section~\ref{sec:failure}, we discuss few examples of failure cases that we mentioned in Section~5 of the main paper.

In Section~\ref{sec:extra-res}, we show more results in addition to the results shown in Fig.~8 of the main paper. In particular, Fig.~\ref{fig:suppl3}$~\sim$~\ref{fig:suppl8} demonstrate the application of our clothes retargeting method to the various clothes template models. These examples highlight the generalization capability of our method.
Finally, for the best visualization of the retargeted 3D mesh (template) onto 2D image, we show several results shown from different viewpoints.

\section{Details on Clothes Retargeting Network}
\label{sec:crn}
In Fig.~\ref{fig:supple_network}, we describe the implementation details of our Clothes Retargeting Network (CRNet). CRNet has four different convolutional blocks. The feature extraction block (blue) has the same structure with VGGNet~\cite{SimonyanZ14a}, where the pre-trained weights are used for initialization. The \textit{compressor} blocks~\cite{shin2017pixel} compress the features in a channel-wise using a convolutional filter with 1$\times$1 resolution and then fully connect to the one dimensional vector (512$\times$1). Note that the layer normalization in the end of each \textit{compressor} block enforces the scale coherency of the features from the different depth-level layers. From the globally combined multiple depth-level vectors (2048$\times$1), deformation regression block (red) and camera pose regression block (green) predict the local deformation field ($\Delta\textbf{M}$) of the clothing model and camera pose $\textbf{C}$, respectively, using several times of the fully connected layers.



\SetKwInput{KwData}{Input}
\SetKwInput{KwSet}{Set}
\SetKwInput{Kwinit}{Initialize}
\SetKwInput{Kwcond}{Function}
\SetKwRepeat{Do}{do}{while}%
\begin{algorithm}[t]
    \KwSet{the learning rate $\eta_{C}=e^{-5}$, $\eta=e^{-5}$, and the weight for each loss term $\lambda_c=1$, $\lambda_d=1$}
    \Kwinit{the pre-trained CRNet weights, $\mathbf{w}_{C}$, $\mathbf{w}$}
    \KwData{the 2D densepose detection $\mathbf{s}$ from real data $\mathcal{I}$.}
    \Kwcond{\textit{feedforward()}\\
    i)\ \ Predict shape and pose: $\Delta\mathbf{M}$, $\mathbf{C}$ $\gets$ CRNet($\mathbf{s}$)\; 
   ii) Compute loss: $\lambda_c\mathcal{L}_c+\lambda_d\mathcal{L}_d$  $\gets$ Eq.~(6, 7)\; 
    }
   \Do{$\phi_C(\mathbf{s};\mathbf{w}_C)$ is not converged}{
   Call \textit{feedforward()}\;
   Update $\mathbf{w}_C$ with $\eta_{C}$, $\lambda_c=1$, $\lambda_d=0$  \;
    }
   \Do{$\phi(\mathbf{s};\mathbf{w})$ is not converged}{
    Call \textit{feedforward()}\;
    Update $\mathbf{w}$ with $\eta$, $\lambda_c=1$, $\lambda_d=0$  \;
    }
    \Do{$\phi_C(\mathbf{s};\mathbf{w}_C)$, $\phi(\mathbf{s};\mathbf{w})$ are not converged}{
      Call \textit{feedforward()}\;
      Update $\mathbf{w}_{C}$, $\mathbf{w}$ with $0.01\cdot\eta_{C}$, $0.1\cdot\eta$, $\lambda_c=0.5$, $\lambda_d=1$ \;  
    }
  \caption{Online Refinement}
  \label{algo}
\end{algorithm}

\section{Online Refinement Algorithm}
\label{sec:algoritm}
As described in Section~3.4, we refine the weight of the pre-traiend CRNet on the testing data with our self-supervision losses. The detailed algorithm flow with the associated parameters are described in Table~\ref{algo}.

\section{Multi-view Data for Evaluations}
\label{sec:multiview}
\begin{figure}[t]
	\begin{center}
		\hspace{-5mm}\includegraphics[width=3.5in]{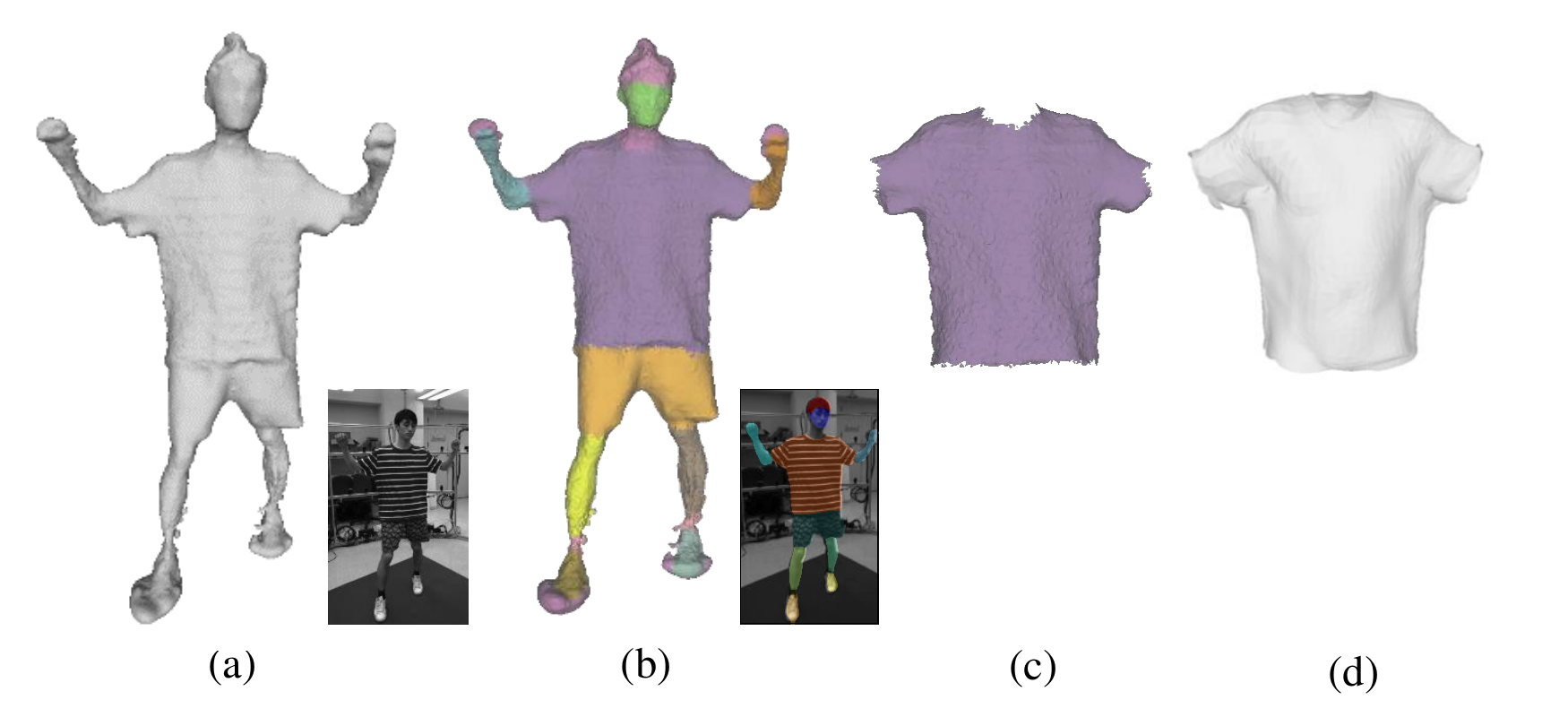}
	\end{center}	
	\vspace{-7mm}
	\caption{The intermediate results of our multiview 3D clothing capturing. (a) 3D reconstruction of a subject using the structure from motion algorithm. (b) 3D semantic segmentation by voting the multiview 2D recognition score, where the color denotes the class of each vertex. (c) Parsing out the mesh which has the class of clothes. (d) Fitting our clothing template model to the 3D clothes reconstruction.}
		\vspace{-1mm}
	\label{fig:supple_multiview}
\end{figure}

As mentioned in Section 4.~in the main paper, we further captured the 3D cloth performance using a large-scale multiple camera system for the quantitative evaluation of our method.
Here we provide a brief overview of our multi-view data capturing process. 
\\
Given a set of multiview images (captured from the synchronized 67 cameras), we first scan a 3D model using standard structure from motion with COLMAP software~\cite{schoenberger2016sfm} (Figure~\ref{fig:supple_multiview} (a)). The 3D model segmentation (Figure~\ref{fig:supple_multiview} (b)) is then performed by voting the 2D recognition score to each 3D vertex as similar to~\cite{shin20183d}, and we then parse out only clothes area as described in Figure~\ref{fig:supple_multiview} (c). We finally fit our cloth template to the reconstructed cloth (Figure~\ref{fig:supple_multiview} (d)) by minimizing the correspondence distance with the regularization constraints~\cite{pons2017clothcap}, where the correspondences are established by fitting SMPL model~\cite{loper2015smpl} to both 3D cloth template and reconstruction.

\begin{figure}[t]
	\begin{center}
		\includegraphics[width=3.2in]{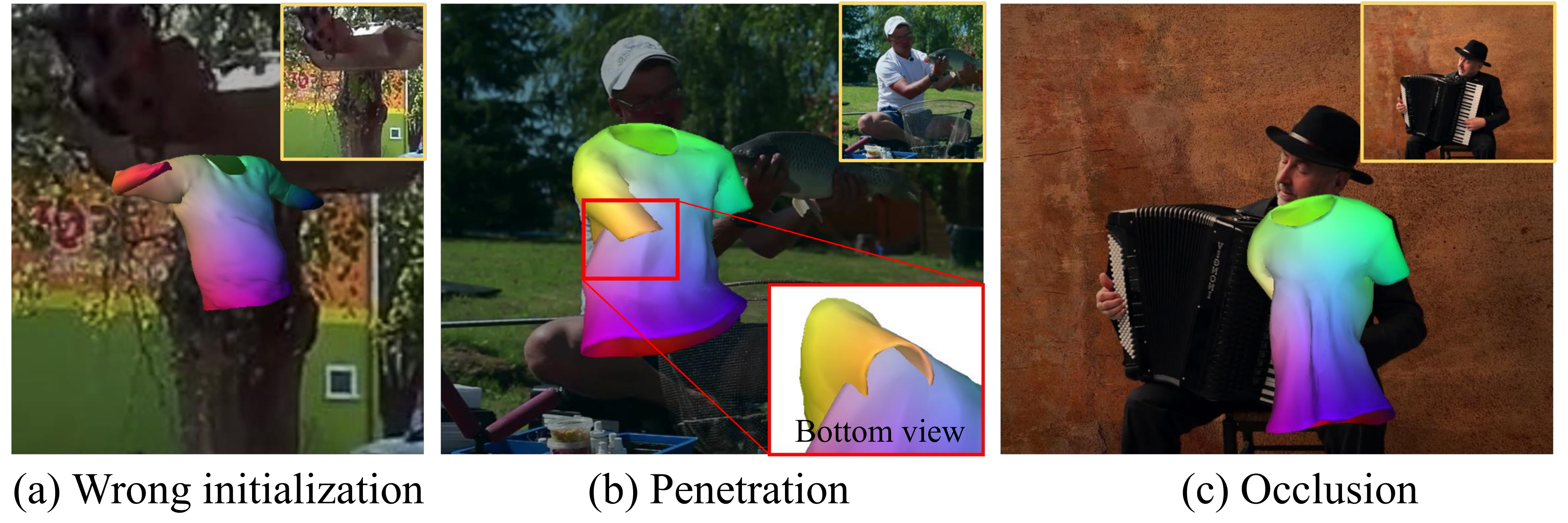}
	\end{center}	
	\vspace{-5mm}
	\caption{The failure cases of our method. The input image is visualized as inset with a yellow bounding box.}
	\label{fig:supple_fail}
		\vspace{-1mm}
\end{figure}

\begin{figure}[t]
	\begin{center}
		\includegraphics[width=0.46\textwidth]{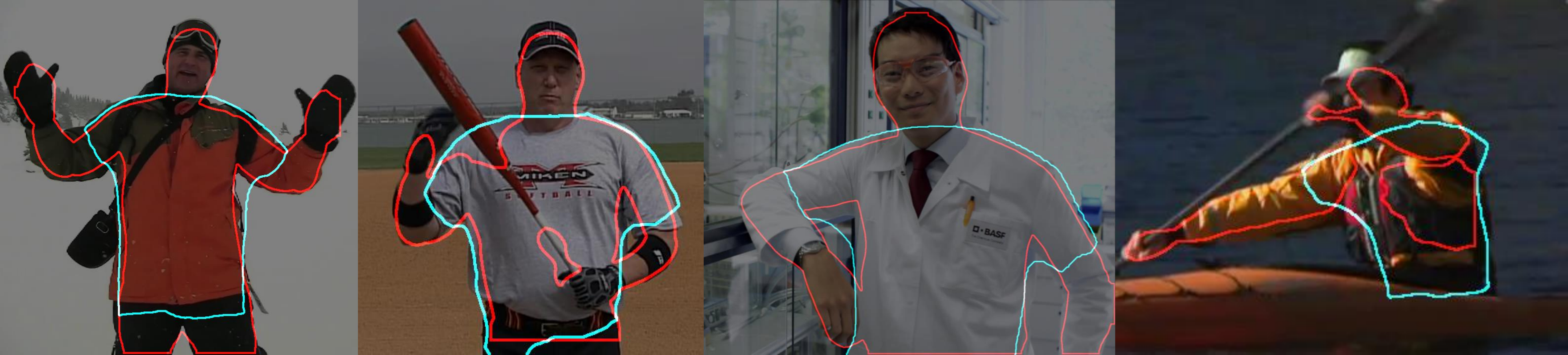}
	\end{center}	
	\vspace{-5mm}
	\caption{\small The qualitative silhouette comparison between the detected densepose mask (red) and the boundary of 3D clothes retargeted by ours (blue). 
	Notice that densepose mainly detects the inside of the body, while CRNet captures semantically more meaningful clothing boundaries. }
	\label{fig:comp_densepose}
	\vspace{-.2cm}
\end{figure}
\begin{figure*}[t]
	\begin{center}
		\includegraphics[width=6.7in]{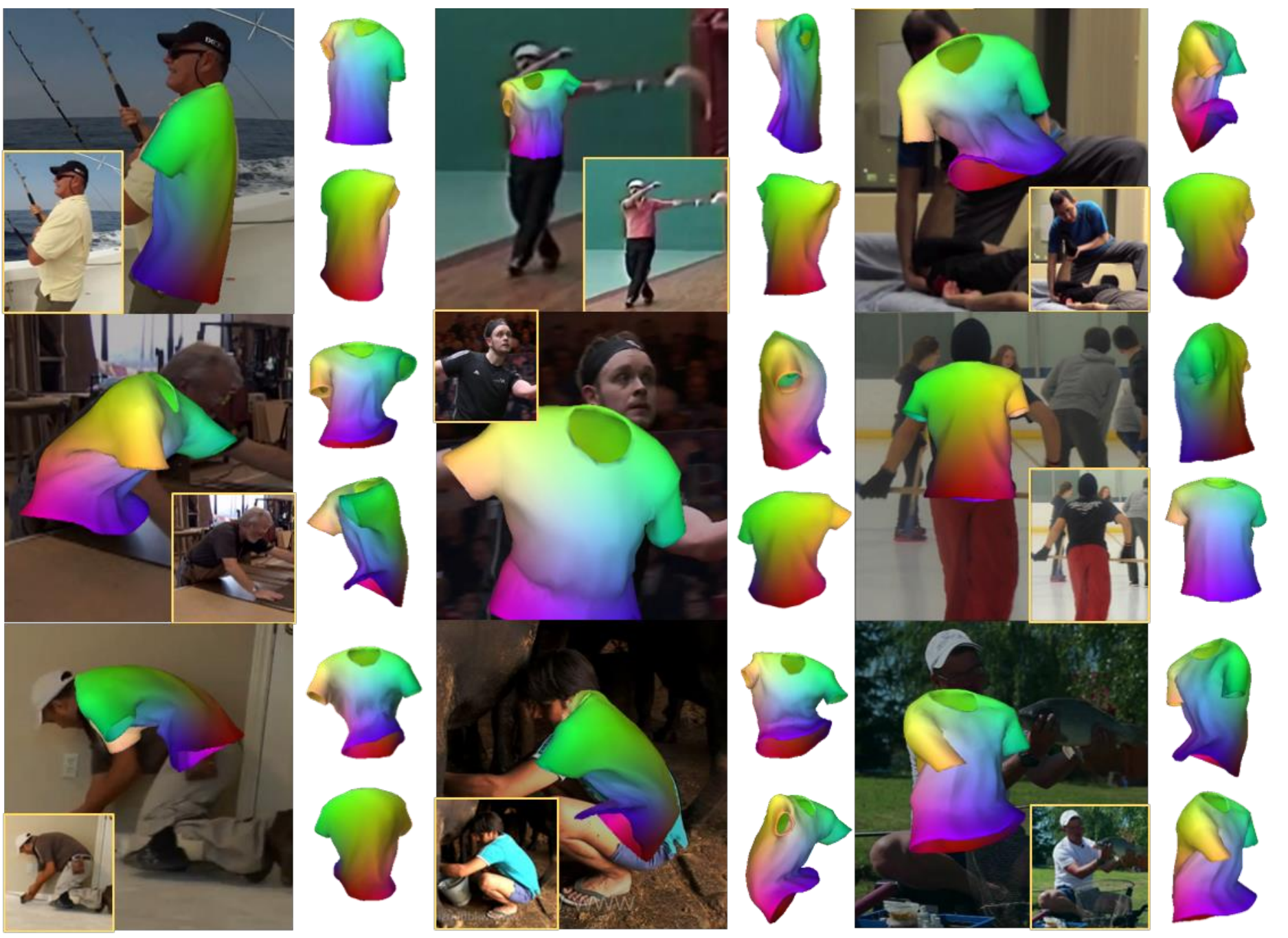}

	\end{center}	
	\caption{The additional clothes retargeting results for half-sleeve shirts model with the shaded surface.} 
	\label{fig:suppl1}
\end{figure*}

\section{Failure Cases}

\label{sec:failure}
Fig.~\ref{fig:supple_fail} shows few examples of the failure cases mentioned in the Section 5~of the main paper. Fig.~\ref{fig:supple_fail} (a) shows the wrong prediction due to the bad initialization from the dense body pose (obviously there is no cloth in the example). 
Fig.~\ref{fig:supple_fail} (b) shows the penetration effect when a part of the cloth has an ambiguity over the other part of the clothing. Fig.~\ref{fig:supple_fail} (c) describes the failure due to severe occlusion mentioned in the main paper as well.

\section{Extra Results}
\label{sec:extra-res}
Fig.~\ref{fig:comp_densepose} shows the comparison of the silhouette detected from densepose with the one from the retargeted 3D clothing. In Fig.~\ref{fig:suppl1}~$\sim$~Fig.~\ref{fig:suppl8}, we provide additional clothes retargeting results with different clothes templates (half sleeve, sleeveless, and long sleeve shirts, and dress model). We show the results with shaded mesh (for clearer visualization of the quality of the retargeted clothes surface) for various input images with people having arbitrary poses and appearances. For all the result images, the input image is added as inset with a yellow bounding box.











\begin{figure*}[t]
	\begin{center}
		\includegraphics[width=6.5in]{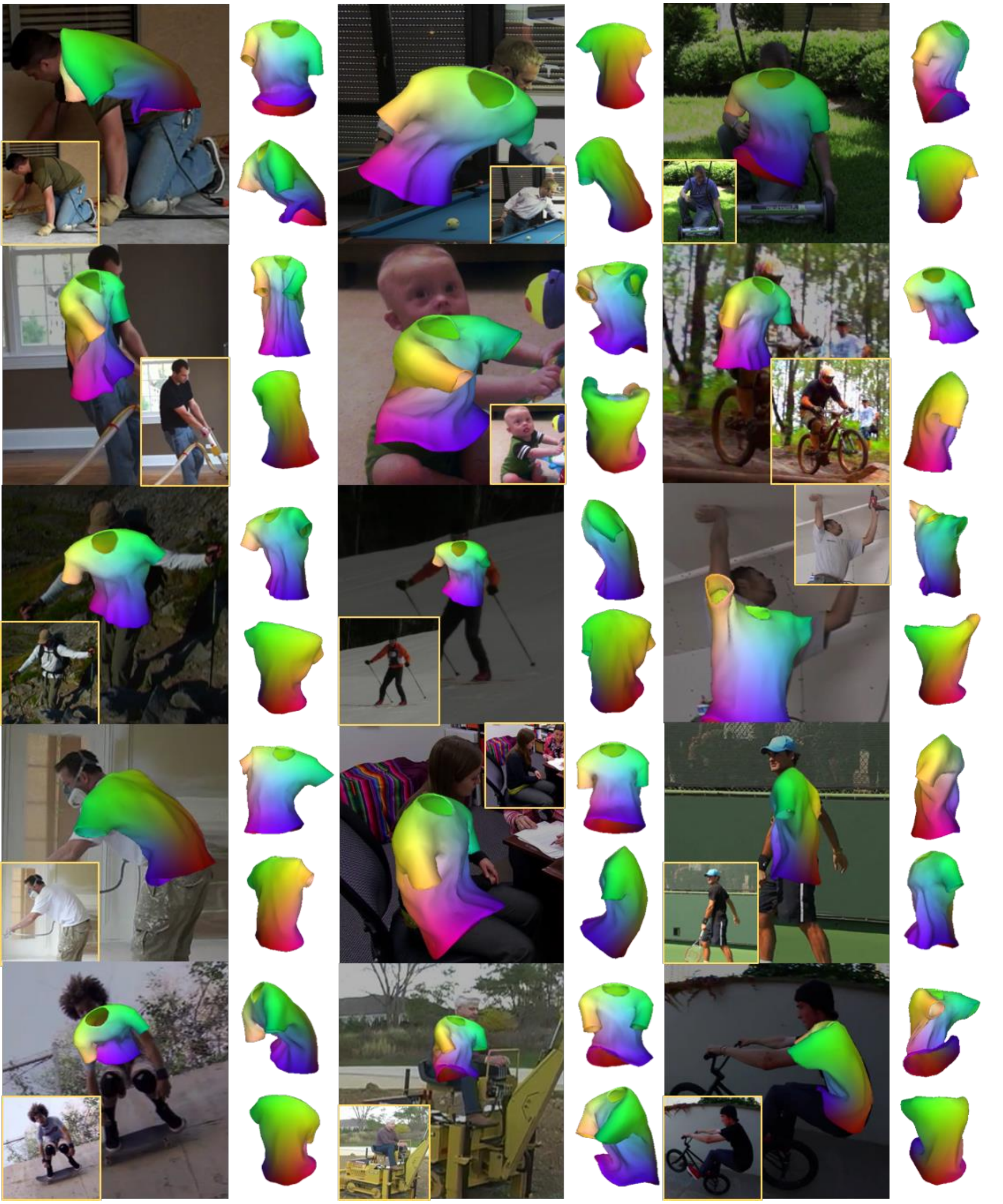}
	\end{center}	
	\vspace{-2mm}
	\caption{The additional clothes retargeting results for half-sleeve shirts model with the shaded surface.} 
	\label{fig:suppl2}
\end{figure*}

\begin{figure*}[t]
	\begin{center}
		\includegraphics[width=6.5in]{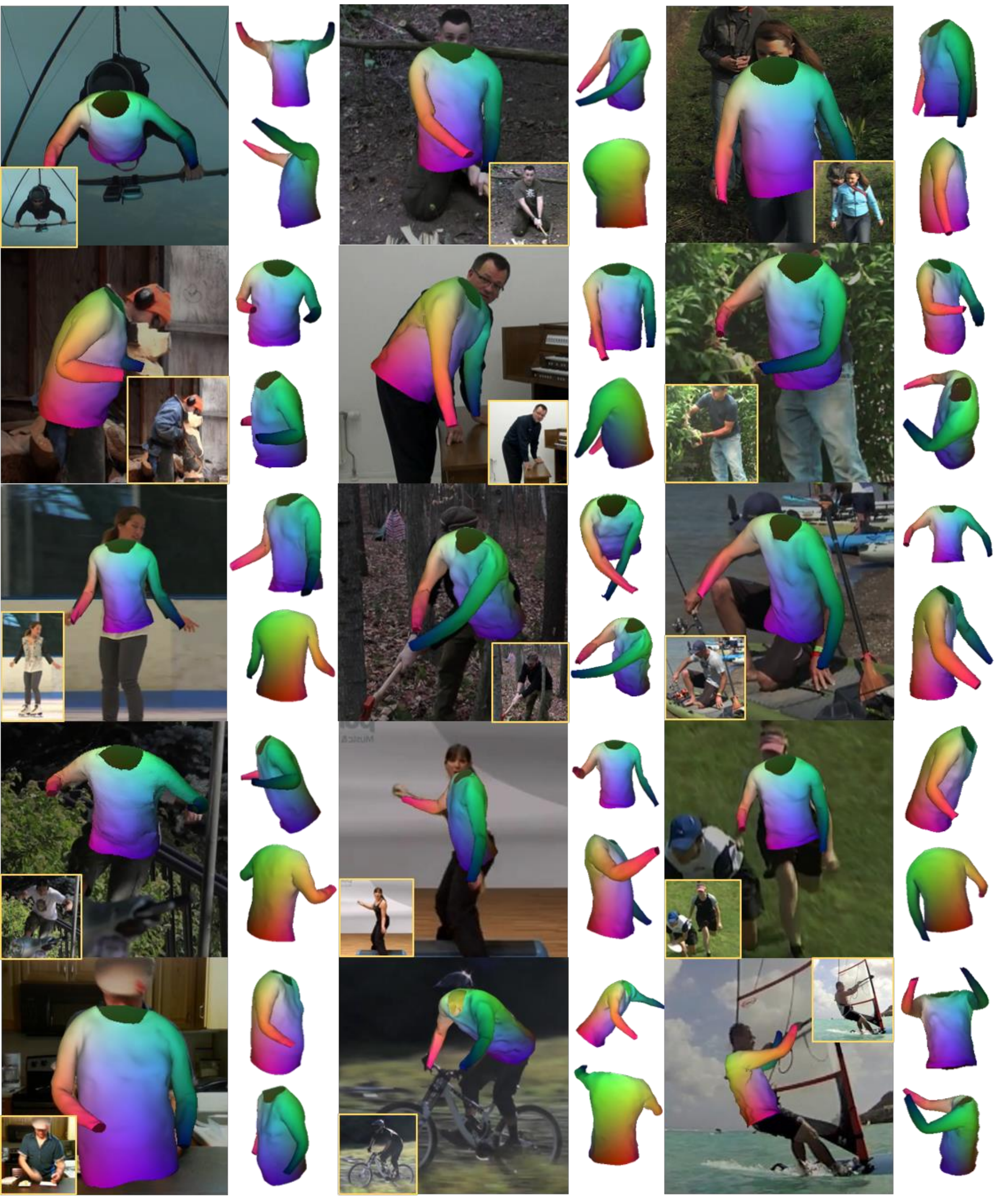}
	\end{center}	
	\vspace{-2mm}
	\caption{The additional clothes retargeting results for long-sleeve shirts model with the shaded surface.} 
	\label{fig:suppl3}
\end{figure*}

\begin{figure*}[t]
	\begin{center}
		\includegraphics[width=6.5in]{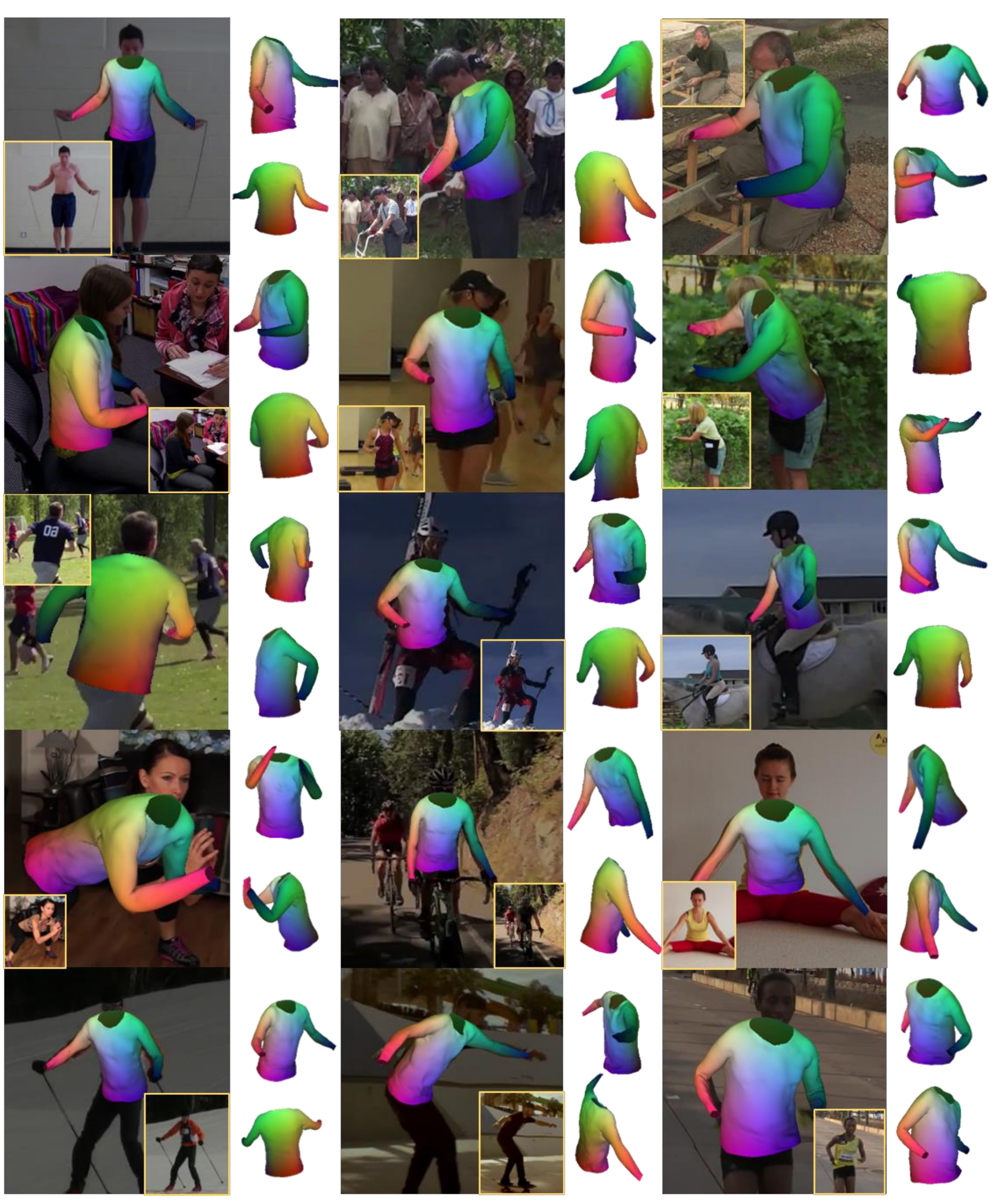}
	\end{center}	
	\vspace{-2mm}
	\caption{The additional clothes retargeting results for long-sleeve shirts model with the shaded surface.} 
	\label{fig:suppl4}
\end{figure*}

\begin{figure*}[t]
	\begin{center}
		\includegraphics[width=6.5in]{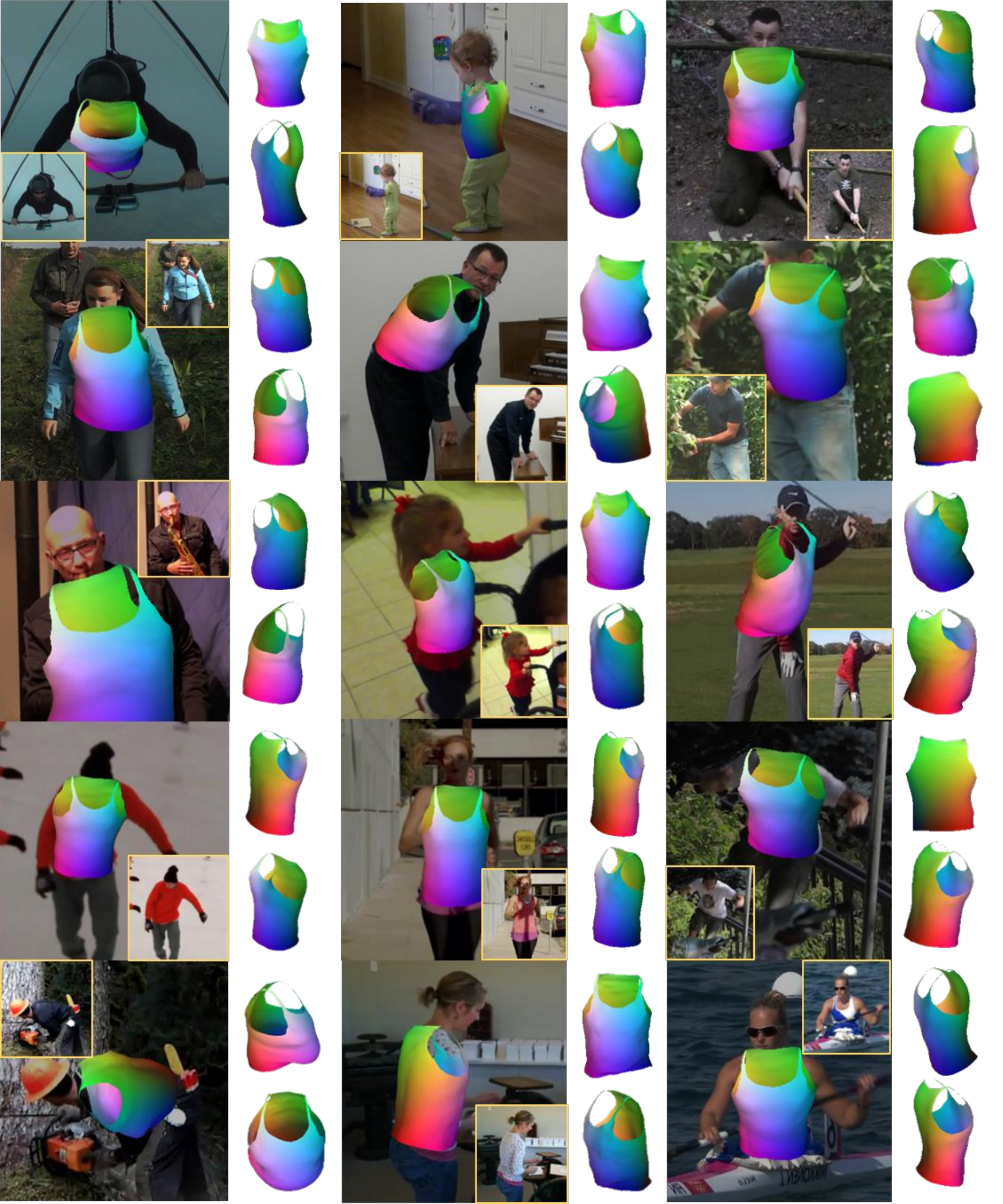}
	\end{center}	
	\vspace{-2mm}
	\caption{The additional clothes retargeting results for sleeveless clothing model with the shaded surface.} 
	\label{fig:suppl7}
\end{figure*}

\begin{figure*}[t]
	\begin{center}
		\includegraphics[width=6.5in]{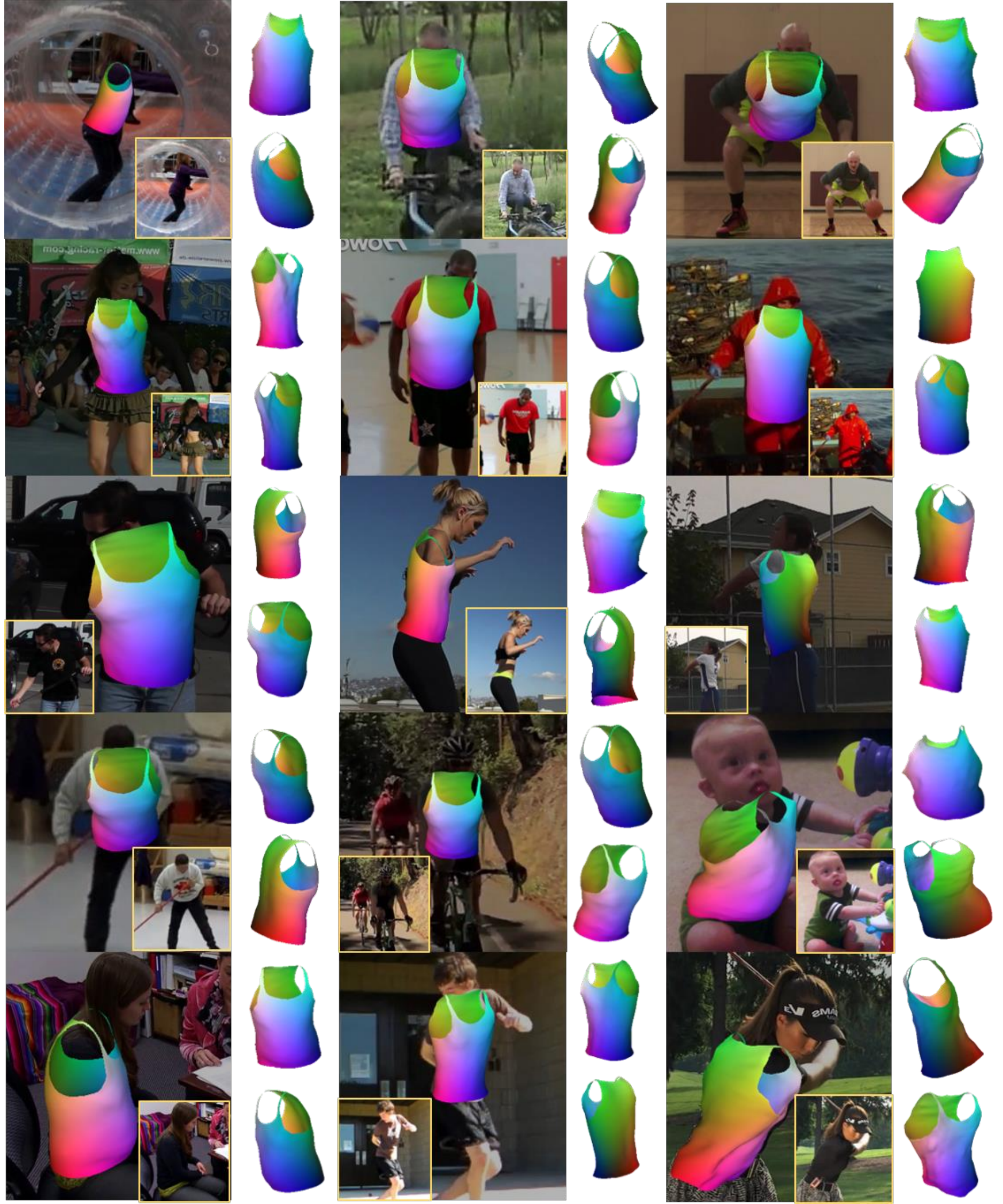}
	\end{center}	
	\vspace{-2mm}
	\caption{The additional clothes retargeting results for sleeveless clothing model with the shaded surface.} 
	\label{fig:suppl8}
\end{figure*}

\begin{figure*}[t]
	\begin{center}
		\includegraphics[width=6.5in]{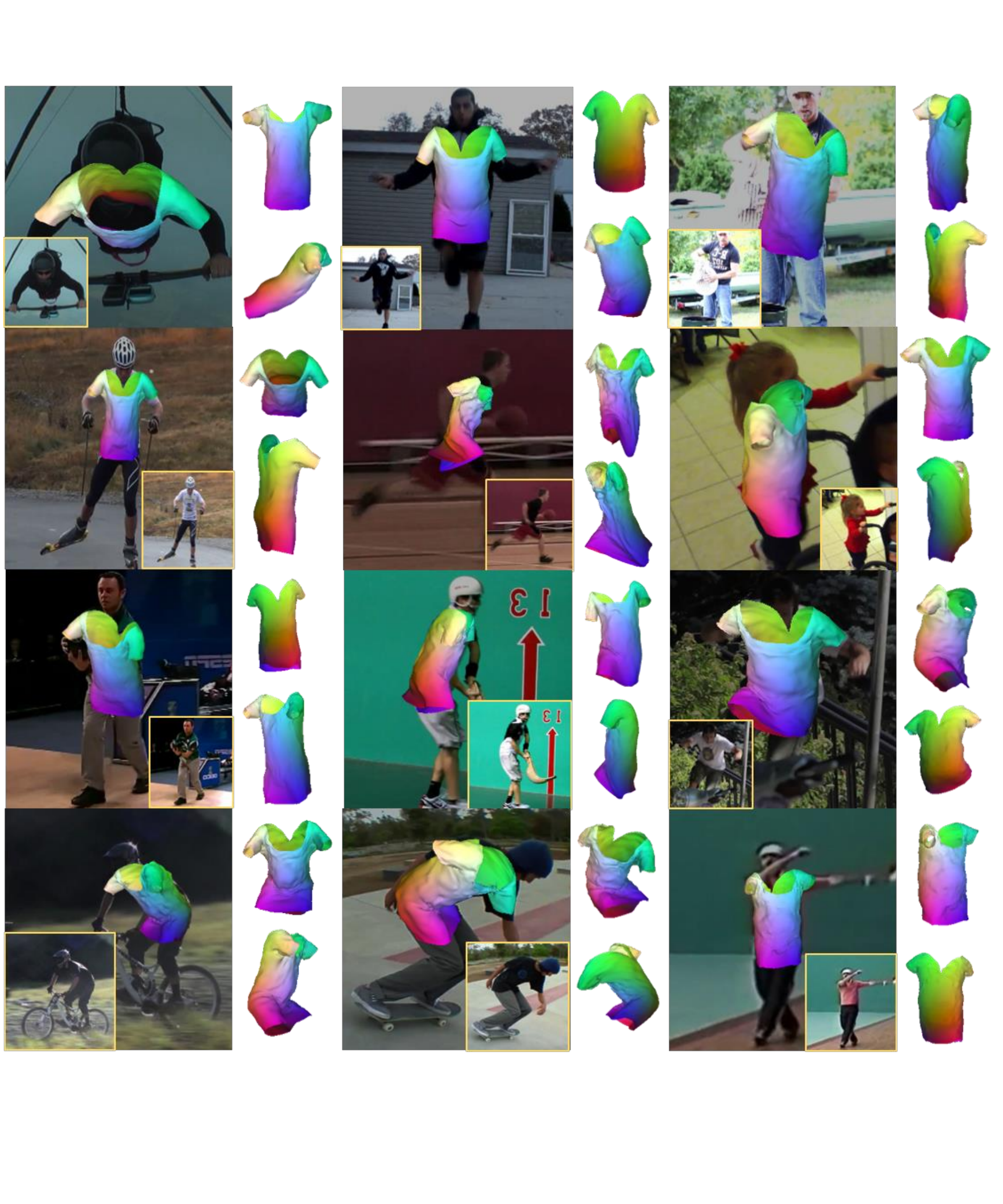}
	\end{center}	
	\vspace{-22mm}
	\caption{The additional clothes retargeting results for dress model with the shaded surface.} 
	\label{fig:suppl5}
\end{figure*}

\begin{figure*}[t]
	\begin{center}
		\includegraphics[width=6.5in]{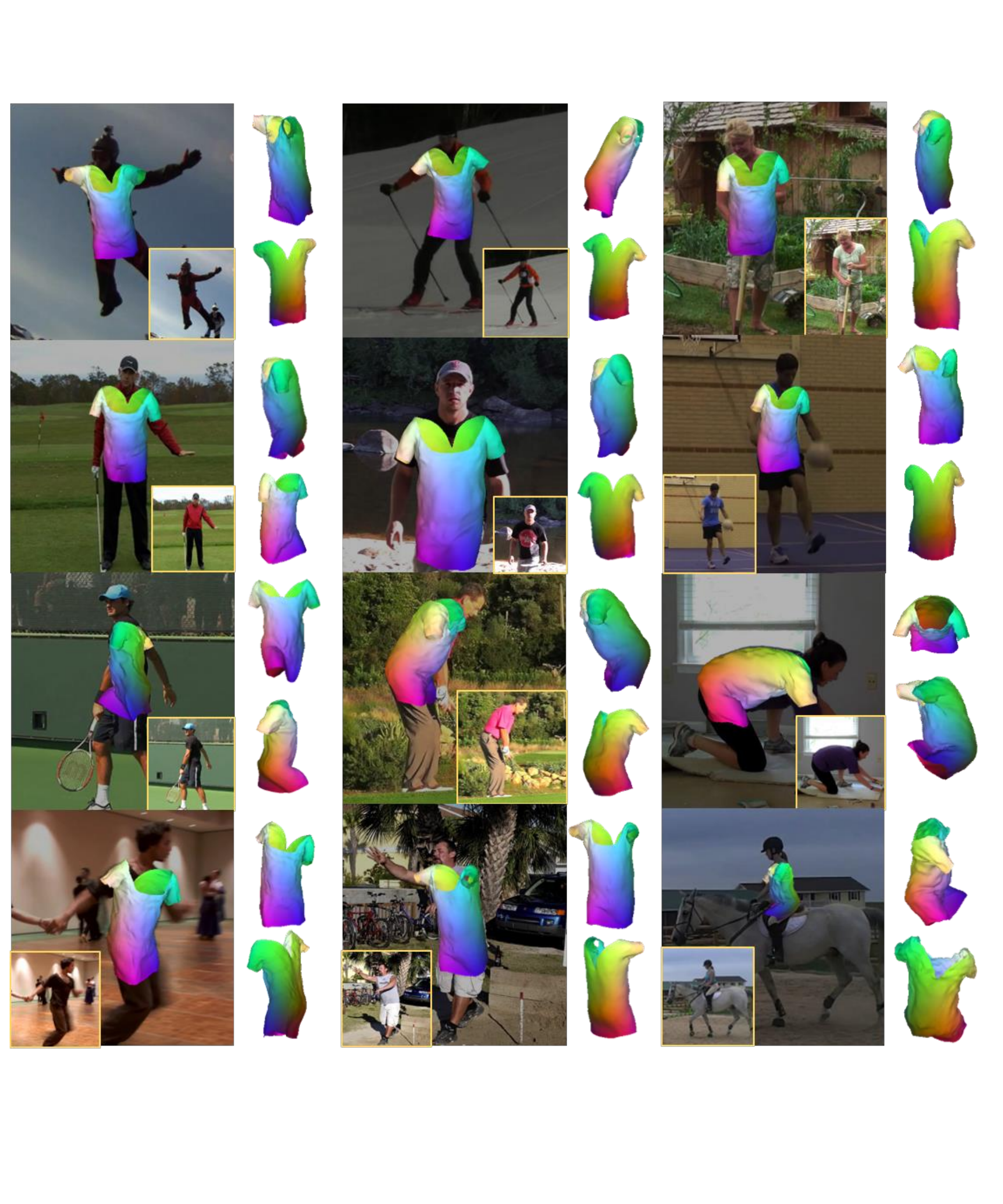}
	\end{center}	
	\vspace{-22mm}
	\caption{The additional clothes retargeting results for dress model with the shaded surface.} 
	\label{fig:suppl6}
\end{figure*}

\end{document}